\documentclass[letterpaper,journal]{IEEEtran}

\usepackage{amsmath,amssymb,amsfonts,amsthm}
\usepackage{graphicx}
\usepackage{booktabs}
\usepackage{multirow}
\usepackage{array}
\usepackage{xcolor}
\usepackage{tikz}
\usetikzlibrary{arrows.meta, positioning, calc, shapes.geometric, fit}
\usepackage[hidelinks]{hyperref}
\usepackage{cite}
\usepackage[final]{microtype}
\DeclareRobustCommand{\_}{\ifmmode\mathchar"5F\else\textunderscore\allowbreak\fi}

\definecolor{cblue}{RGB}{31,119,180}
\definecolor{cred}{RGB}{214,39,40}
\definecolor{cgreen}{RGB}{44,160,44}
\definecolor{corange}{RGB}{230,115,0}
\definecolor{cpurple}{RGB}{148,103,189}
\definecolor{cgray}{RGB}{110,110,110}

\newtheorem{proposition}{Proposition}

\newcommand{\MsPreRegret}{0.040}

\newcommand{\MsPreMapAde}{20.0}

\newcommand{\MsEteRegret}{0.048}
\newcommand{\MsEteMapAde}{20.7}
\newcommand{\MsRawSoftRegret}{0.042}
\newcommand{\MsRawSoftMapAde}{20.9}
\newcommand{\MsRawSoftRbErr}{9.14}
\newcommand{\MsPureCeRegret}{0.042}

\newcommand{\MsPureCeMapAde}{25.6}

\newcommand{\MsPureCeKabRegret}{0.044}
\newcommand{\MsPureCeKabMapAde}{23.2}
\newcommand{\MsSeTwoSoftRegret}{0.049}
\newcommand{\MsSeTwoSoftMapAde}{22.9}

\newcommand{\MsArRegret}{0.076}

\newcommand{\PubEteMapAde}{10.17}

\newcommand{\PubSoftRawMapAde}{12.78}
\newcommand{\PubSoftKabMapAde}{11.76}
\newcommand{\PubSeTwoSoftMapAde}{13.14}
\newcommand{\PubPureCeRawMapAde}{16.53}
\newcommand{\PubPureCeKabMapAde}{14.38}

\newcommand{\PubArRawMapAde}{25.43}
\newcommand{\PubArKabMapAde}{24.11}
\newcommand{\PubBiFpdRawMapAde}{16.75}
\newcommand{\PubBiFpdKabMapAde}{15.79}
\newcommand{\BfMsEteRegret}{0.020}
\newcommand{\BfMsEteRegretCI}{[0.016,\,0.025]}
\newcommand{\BfMsEteTopOne}{73.2}
\newcommand{\BfMsEteMapAde}{11.35}
\newcommand{\BfMsPreRegret}{0.022}
\newcommand{\BfMsPreRegretCI}{[0.018,\,0.027]}

\newcommand{\BfMsPreMapAde}{11.42}

\newcommand{\BfMsPureCeKabRegret}{0.021}
\newcommand{\BfMsPureCeKabMapAde}{11.70}

\newcommand{\BfPubEteMapAde}{3.78}
\newcommand{\BfPubOrthoEteMapAde}{3.72}

\newcommand{\WfNoRgbMsRegret}{0.043}
\newcommand{\WfNoRgbMsMapAde}{19.31}
\newcommand{\BfRotRgbMsEteRegret}{0.028}
\newcommand{\BfRotRgbMsEteMapAde}{13.78}
\newcommand{\WfNoRgbPubKabMapAde}{11.84}

\newcommand{\ci}[1]{\,\text{\footnotesize$#1$}}

\newcommand{\pholm}[1]{p_{\text{Holm}(#1)}}
\newcommand{\SE}{\text{SE}(2)}

\begin{document}

\title{Frame First, Decode Once: Object-Centric Coordinates and Single-Pass Decoding for System-1 World Models}

\author{Abu~Anas~Ibn~Samad%
\thanks{The author is with Sigmind.ai, Dhaka 1215, Bangladesh (e-mail: \mbox{abushuvom@sigmind.ai}).}}

\markboth{Preprint, November 2026}%
{Samad: Frame First, Decode Once}

\maketitle

\begin{abstract}
Single-pass primitive decoders such as LocateAnything and VFlash predict all quantized coordinate tokens of a geometric primitive in one forward pass, enabling fast, System-1 visual planning. We study what matters when such a decoder serves as a System-1 world model: a fast action-conditioned predictor that ranks candidate pushes by their predicted outcome. On ManiSkill3 \texttt{PushT-v1} ($512$ held-out test families, $8{,}192$ rollouts) and on human demonstrations from \texttt{lerobot/pusht} ($2{,}384$ test windows), we compare decoders, training losses, rigid-body projections and coordinate frames, using matched checkpoints, hierarchical bootstrap confidence intervals and Holm correction. Two factors of similar size dominate. First, expressing inputs and outputs in the $\SE$ frame of the object's initial pose lowers decision regret by $1.9\times$--$2.2\times$ for each of five decoders ($\pholm{9}{<}0.005$), and the object-frame model trained on $128$ families matches the world-frame model trained on $512$. Second, single-pass decoding outperforms iterative decoding: in the object frame, single-pass heads reach $0.020$--$0.021$ regret at $1.6$--$1.8\,\text{ms}$ per decision, against $0.046$ for 10-step DDIM diffusion and $0.059$ for KV-cached autoregression at $10.3$ and $12.9\,\text{ms}$. The choice among single-pass heads has no measurable effect on regret, and our registered hypothesis that the frame effect exceeds the decoder spread is not supported ($0.0345$ vs.\ $0.0389$ regret). Soft-token expectation with an end-to-end 2D Kabsch projection gives the lowest trajectory error ($11.35\,\text{px}$ on ManiSkill3, $3.78\,\text{px}$ on \texttt{lerobot/pusht}), exact rigid-body consistency, and lower regret than diffusion at every budget from $4$ to $50$ DDIM steps, at $3\times$--$29\times$ lower latency.
\end{abstract}

\begin{IEEEkeywords}
World models, robotic manipulation, parallel decoding, coordinate frames, $\SE$ equivariance, orthogonal Procrustes, diffusion models.
\end{IEEEkeywords}

\section{Introduction}
\IEEEPARstart{A}{ction-conditioned} world models let an agent evaluate candidate actions by predicting their outcomes before acting~\cite{chi2023diffusion,ha2018world,hafner2023mastering}. A model that is queried inside the control loop, a fast ``System-1'' model in the sense of Kahneman~\cite{kahneman2011thinking}, must do two things: run within a few milliseconds, and rank the candidate actions correctly. Vision-language and vision-language-action models represent geometry as sequences of quantized coordinate tokens $\langle 0\rangle\dots\langle 1000\rangle$~\cite{rt2,openvla,qwen25vl}. Decoding such a trajectory token by token takes one sequential step per token, i.e.\ $P \times C$ steps for $P$ time steps of $C$ coordinates ($100$ for $P{=}C{=}10$). Even a coarser autoregression that emits one time step at a time, as in our baseline, needs about $12\,\text{ms}$ per decision for a $3.8$M-parameter model with a key-value cache, against about $1\,\text{ms}$ for a single forward pass.

LocateAnything~\cite{locateanything} and VFlash~\cite{vflash} remove this bottleneck. Once the type of a primitive is known, its coordinate layout is fixed, so all of its coordinate slots can be predicted jointly in a single forward pass: slot positions attend to each other and to the preceding context. VFlash applies this flash primitive decoding (FPD) to visual plans such as robot trajectories and driving waypoints, and trains the slots with token-level cross-entropy.

We study single-pass slot decoders used as System-1 world models for contact-rich manipulation. The model predicts the trajectory of a pushed T-block under each of four candidate actions, and the agent picks the action with the highest predicted utility. Prior work leaves several design choices open, which we phrase as five questions:
\begin{enumerate}
\item How does single-pass decoding compare with iterative decoders (DDIM diffusion, autoregression) built on the same backbone and data?
\item Does supervising a soft-token expectation with a metric loss improve on token cross-entropy?
\item Should rigid-body consistency come from direct $\SE$ pose tokens or from redundant keypoints followed by a Procrustes projection?
\item Should the projection be applied only at test time or trained end to end?
\item Which matters more for decision quality, the coordinate frame or the decoder?
\end{enumerate}

We answer these questions on ManiSkill3 \texttt{PushT-v1}~\cite{maniskill3} and on the human demonstrations of \texttt{lerobot/pusht}~\cite{chi2023diffusion,lerobot}. The study has three stages. A benchmark of twelve models was pre-registered before its test split was generated. Exploratory controls followed, including the object-frame models. Finally, a $2{\times}5$ frame $\times$ decoder grid was registered after the exploratory object-frame result and run as a confirmatory replication on the same test families. Projected and unprojected variants of each head are evaluated on the same checkpoints, and each table of paired tests applies one Holm correction across all competitors it reports.

Our findings are:
\begin{itemize}
\item \textbf{The coordinate frame.} Expressing inputs and outputs in the $\SE$ frame of the object's initial pose halves decision regret for every decoder we tested and gives roughly $4\times$ sample efficiency. Controls on the RGB input show that the gain comes from the coordinate frame and not from removing the image.
\item \textbf{Single-pass versus iterative decoding.} On the same backbone, single-pass heads have $2.3\times$--$2.9\times$ lower regret than a 10-step DDIM diffusion model and a KV-cached autoregressive model, and are $6\times$--$8\times$ faster. Which single-pass head is used makes no measurable difference to regret. The frame effect and the decoder effect are of comparable size, so our registered hypothesis that the frame dominates is not supported.
\item \textbf{Procrustes-FPD.} Soft-token expectation followed by a closed-form 2D Kabsch projection gives exact rigid-body consistency and the lowest trajectory error. Redundant keypoints with Procrustes fitting outperform direct $\SE$ tokens in the world frame. We characterize when to apply the projection at test time and when to train through it, and we relate this trade-off to the rank of the projection's Jacobian.
\end{itemize}
We also report an implementation error in our original diffusion baseline, its correction, and how it changed the results (Section~\ref{sec:grid_scaling}, Appendix~\ref{app:superseded}).

Throughout, ``VFlash-style CE head'' denotes our reimplementation of VFlash's slot loss and decoding inside the same $2$M-parameter backbone as every other model. We do not evaluate NVIDIA's released model, and our results concern design choices for small action-conditioned predictors, not the VFlash system.

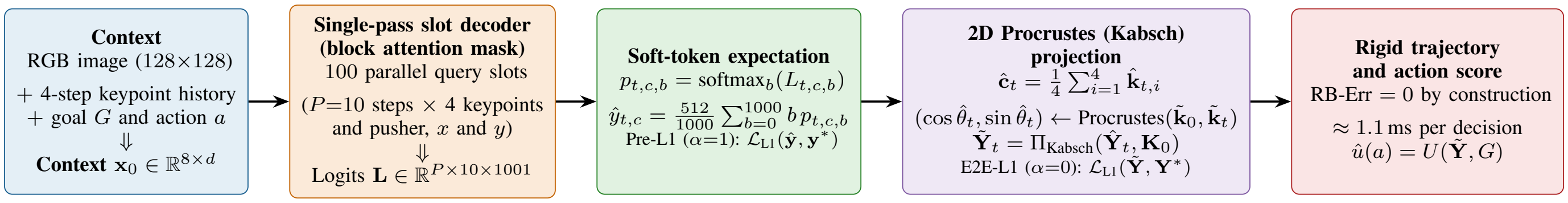
\begin{figure*}[t]
\centering
\resizebox{\textwidth}{!}{%
\begin{tikzpicture}[
    font=\footnotesize,
    node distance=0.45cm and 0.55cm,
    box/.style={draw, rounded corners=3pt, align=center, minimum height=0.85cm, inner sep=4.5pt},
    ctx/.style={box, fill=cblue!12, draw=cblue!80!black},
    slotbox/.style={box, fill=corange!15, draw=corange!85!black},
    softbox/.style={box, fill=cgreen!14, draw=cgreen!80!black},
    kabschbox/.style={box, fill=cpurple!15, draw=cpurple!85!black},
    outbox/.style={box, fill=cred!12, draw=cred!80!black},
    arr/.style={-{Stealth[length=2.2mm]}, thick}
]
\node[ctx, minimum width=3.1cm, minimum height=2.5cm] (ctx) {
    \textbf{Context}\\
    \vspace{2pt}
    RGB image ($128{\times}128$)\\
    $+$ 4-step keypoint history\\
    $+$ goal $G$ and action $a$\\
    $\Downarrow$\\
    \textbf{Context $\mathbf{x}_0 \in \mathbb{R}^{8 \times d}$}
};

\node[slotbox, right=0.55cm of ctx, minimum width=3.3cm, minimum height=2.5cm] (sdpa) {
    \textbf{Single-pass slot decoder}\\
    \textbf{(block attention mask)}\\
    \vspace{2pt}
    $100$ parallel query slots\\
    ($P{=}10$ steps $\times$ 4 keypoints\\
    and pusher, $x$ and $y$)\\
    $\Downarrow$\\
    Logits $\mathbf{L} \in \mathbb{R}^{P \times 10 \times 1001}$
};

\node[softbox, right=0.55cm of sdpa, minimum width=3.4cm, minimum height=2.5cm] (soft) {
    \textbf{Soft-token expectation}\\
    \vspace{2pt}
    $p_{t,c,b} = \text{softmax}_b(L_{t,c,b})$\\
    $\hat{y}_{t,c} = \frac{512}{1000}\sum_{b=0}^{1000} b\,p_{t,c,b}$\\
    \vspace{2pt}
    {\scriptsize Pre-L1 ($\alpha{=}1$): $\mathcal{L}_{\text{L1}}(\hat{\mathbf{y}}, \mathbf{y}^*)$}
};

\node[kabschbox, right=0.55cm of soft, minimum width=3.5cm, minimum height=2.5cm] (kab) {
    \textbf{2D Procrustes (Kabsch)}\\
    \textbf{projection}\\
    \vspace{2pt}
    $\hat{\mathbf{c}}_t = \frac{1}{4}\sum_{i=1}^4 \hat{\mathbf{k}}_{t,i}$\\
    $(\cos\hat{\theta}_t, \sin\hat{\theta}_t) \leftarrow \text{Procrustes}(\tilde{\mathbf{k}}_0, \tilde{\mathbf{k}}_t)$\\
    $\tilde{\mathbf{Y}}_t = \Pi_{\text{Kabsch}}(\hat{\mathbf{Y}}_t, \mathbf{K}_0)$\\
    {\scriptsize E2E-L1 ($\alpha{=}0$): $\mathcal{L}_{\text{L1}}(\tilde{\mathbf{Y}}, \mathbf{Y}^*)$}
};

\node[outbox, right=0.55cm of kab, minimum width=2.4cm, minimum height=2.5cm] (out) {
    \textbf{Rigid trajectory}\\
    \textbf{and action score}\\
    \vspace{2pt}
    RB-Err $= 0$ by construction\\
    $\approx 1.1\,\text{ms}$ per decision\\
    $\hat{u}(a) = U(\tilde{\mathbf{Y}}, G)$
};

\draw[arr] (ctx) -- (sdpa);
\draw[arr] (sdpa) -- (soft);
\draw[arr] (soft) -- (kab);
\draw[arr] (kab) -- (out);
\end{tikzpicture}}
\caption{Procrustes-FPD. Given a context $\mathbf{x}_0$ and a candidate action $a$, a 3-layer transformer ($d{=}192$) with a block attention mask predicts the token distributions of all $100$ trajectory coordinates in one forward pass. The soft-token expectations of the four object keypoints, $\hat{\mathbf{Y}}_t \in \mathbb{R}^{4\times 2}$, are projected onto the rigid motions of the observed initial shape $\mathbf{K}_0$ by a closed-form 2D Procrustes fit, $\Pi_{\text{Kabsch}}$. The projection is applied either at test time only (Pre-L1, $\alpha{=}1$) or inside the training loss (E2E-L1, $\alpha{=}0$). The predicted trajectory is scored with the task utility $U$. Latency is for one decision over four candidate actions (Table~\ref{tab:maniskill_main}).}
\label{fig:architecture}
\end{figure*}

\section{Related Work}
\subsection{Parallel Primitive Decoding}
LocateAnything~\cite{locateanything} decodes each bounding box or point of a grounding answer as one unit in a single step, and VFlash~\cite{vflash} extends this idea to visual plans on top of a speculative drafter. Given the context of a frozen vision-language model, VFlash appends the $B$ positions of a primitive block. Positions inside the block attend to each other and to the context before the block, and all are predicted in one forward pass. Primitive positions are trained with token-level cross-entropy (Eq.~4 of~\cite{vflash}, with unit weights on primitive positions):
\begin{equation}
\label{eq:vflash_ce}
\mathcal{L}_{\text{CE}} = -\frac{1}{P C}\sum_{t=1}^{P}\sum_{c=1}^{C} \log P_{\theta}\!\left(z^*_{t,c} \mid \mathbf{x}_0, a\right),
\end{equation}
where, in our setting, $z^*_{t,c} = \text{round}(1000 \cdot y^*_{t,c} / 512) \in \{0,\dots,1000\}$. Eq.~\eqref{eq:vflash_ce} removes the latency of autoregressive decoding. It is, however, indifferent to the distance between bins: predicting $\langle 501\rangle$ or $\langle 999\rangle$ for a target $\langle 500\rangle$ incurs the same loss as long as both receive the same probability.

\subsection{Soft-Bin Regression and Keypoint-Based Pose Estimation}
Taking the expectation over discrete bins so that a continuous metric loss can be applied was introduced in integral human pose regression~\cite{sun2018integral}, differentiable spatial-to-numerical transforms~\cite{nibali2018numerical}, and Distribution Focal Loss for bounding-box edges~\cite{li2020generalized}. In 6-DoF pose estimation and point-cloud registration, predicting redundant keypoints and fitting the pose in closed form with Kabsch~\cite{kabsch1976solution} or orthogonal Procrustes~\cite{schonemann1966generalized} is known to outperform direct regression of angles or quaternions~\cite{peng2019pvnet,zhou2019continuity,choy2020deep,bregier2021deep}. We combine the two ideas inside a single-pass slot decoder and analyze the difference between applying the projection at test time and training through it.

\subsection{Object-Centric and Equivariant Coordinate Frames}
Expressing states and actions in an agent- or object-centric frame is standard in motion forecasting, e.g.\ VectorNet~\cite{gao2020vectornet} and MTR~\cite{shi2022motion}, and equivariant architectures exploit the same symmetry in manipulation, e.g.\ Equivariant Diffusion Policy~\cite{wang2024equivariant} and EquiBot~\cite{yang2024equibot}. Tokenized decoders, in contrast, usually express coordinates in the frame of the sensor: VFlash normalizes image coordinates to $[0, 1000]$ and places 3D robot coordinates in the current camera frame (and driving waypoints in the ego frame), and general-purpose VLA models also tokenize image or robot-base coordinates~\cite{rt2,openvla,qwen25vl}. We measure how much the frame matters relative to the choice of decoder, for five decoder families, and examine where the gain comes from.

\section{Method}
\label{sec:methodology}

\subsection{Problem Formulation}
At decision time $t_0$ the world model receives an RGB image $I_{t_0} \in [0,1]^{3 \times 128 \times 128}$, a 4-step history $\mathbf{H}_{t_0-3:t_0} \in \mathbb{R}^{4 \times 10}$ of the $K{=}4$ object keypoints $\mathbf{K}_t \in \mathbb{R}^{4 \times 2}$ and the pusher position $\mathbf{e}_t \in \mathbb{R}^2$, the goal keypoints $\mathbf{G} \in \mathbb{R}^{4 \times 2}$, and a candidate action $\mathbf{a} \in \mathbb{R}^4$. It predicts the $P{=}10$-step future $\mathbf{Y}_{1:P} = \{(\mathbf{K}_t, \mathbf{e}_t)\}_{t=1}^{P} \in [0,512]^{P \times 10}$. Each candidate is scored with the task utility $U(\mathbf{Y}_{1:P}, \mathbf{G}, \mathbf{K}_0) \in [0,1]$, which rewards final keypoints close to the goal and penalizes the final orientation error (Section~\ref{sec:metrics}), and the agent selects the candidate with the highest predicted utility.

\subsection{Soft-Token Expectation}
\label{sec:method_softl1}
Let $\mathbf{L}_{t,c} \in \mathbb{R}^{1001}$ be the logits of slot $(t,c)$, for time step $t \in \{1,\dots,P\}$ and coordinate $c \in \{1,\dots,10\}$, and $p_{t,c,b} = \text{softmax}_b(L_{t,c,b})$. Instead of decoding the most likely token, as the CE head does, we take the expectation~\cite{sun2018integral,li2020generalized} in workspace units:
\begin{equation}
\label{eq:soft_expectation}
\hat{y}_{t,c} = \frac{512}{1000}\sum_{b=0}^{1000} b \cdot p_{t,c,b}.
\end{equation}
Its gradient with respect to each logit is proportional to the distance between that bin and the current estimate,
\begin{equation}
\frac{\partial \hat{y}_{t,c}}{\partial L_{t,c,b}} = \frac{512}{1000}\,p_{t,c,b}\!\left(b - \frac{1000}{512}\hat{y}_{t,c}\right),
\end{equation}
so a metric loss on $\hat{y}_{t,c}$ penalizes far-away bins more than near ones.

\subsection{2D Procrustes (Kabsch) Projection}
\label{sec:method_procrustes}
Even with a metric loss, the four keypoints are predicted independently in $\mathbb{R}^8$. Small inconsistencies between their predicted motions deform the block: the Soft-L1 head has a mean pairwise-distance error of \MsRawSoftRbErr\,px on ManiSkill3, and every predicted trajectory violates rigidity (Table~\ref{tab:maniskill_main}).

To enforce rigidity, let $\mathbf{K}_0 = \{\mathbf{k}_{0,i}\}_{i=1}^{K}$ be the observed initial keypoints with centroid $\bar{\mathbf{k}}_0$ and centered coordinates $\tilde{\mathbf{k}}_{0,i} = \mathbf{k}_{0,i} - \bar{\mathbf{k}}_0$. For the predicted keypoints $\hat{\mathbf{K}}_t = \{\hat{\mathbf{k}}_{t,i}\}_{i=1}^{K}$ at step $t$ we find the rigid motion that best explains them:
\begin{equation}
\label{eq:procrustes_obj}
(\hat{\mathbf{R}}_t, \hat{\mathbf{c}}_t) = \underset{\mathbf{R} \in \text{SO}(2),\,\mathbf{c} \in \mathbb{R}^2}{\arg\min} \sum_{i=1}^{K} \left\|\mathbf{c} + \mathbf{R}\tilde{\mathbf{k}}_{0,i} - \hat{\mathbf{k}}_{t,i}\right\|_2^2.
\end{equation}
In 2D the minimizer has a closed form that needs neither an SVD nor an $\text{atan2}$~\cite{kabsch1976solution,schonemann1966generalized}:
\begin{align}
\hat{\mathbf{c}}_t &= \frac{1}{K}\sum_{i=1}^{K}\hat{\mathbf{k}}_{t,i}, \qquad \tilde{\mathbf{k}}_{t,i} = \hat{\mathbf{k}}_{t,i} - \hat{\mathbf{c}}_t, \label{eq:kabsch_centroid}\\
\alpha_t &= \sum_{i=1}^{K}\left(\tilde{k}_{0,i,x}\tilde{k}_{t,i,x} + \tilde{k}_{0,i,y}\tilde{k}_{t,i,y}\right), \label{eq:kabsch_cos}\\
\beta_t &= \sum_{i=1}^{K}\left(\tilde{k}_{0,i,x}\tilde{k}_{t,i,y} - \tilde{k}_{0,i,y}\tilde{k}_{t,i,x}\right), \label{eq:kabsch_sin}\\
\cos\hat{\theta}_t &= \frac{\alpha_t}{\sqrt{\alpha_t^2 + \beta_t^2 + \epsilon}}, \quad \sin\hat{\theta}_t = \frac{\beta_t}{\sqrt{\alpha_t^2 + \beta_t^2 + \epsilon}}. \label{eq:kabsch_norm}
\end{align}
The projected keypoints are $\Pi_{\text{Kabsch}}(\hat{\mathbf{K}}_t, \mathbf{K}_0)_i = \hat{\mathbf{c}}_t + \hat{\mathbf{R}}(\hat{\theta}_t)\tilde{\mathbf{k}}_{0,i}$. The map is differentiable wherever $\alpha_t^2 + \beta_t^2 > 0$.

\textbf{Why keypoints rather than pose tokens.} Suppose each keypoint slot had independent, isotropic error with variance $\sigma^2$ per coordinate. Linearizing~\eqref{eq:kabsch_centroid}--\eqref{eq:kabsch_norm} around the true pose shows that the projection keeps only the component of the $2K$-dimensional error that lies in the 3-dimensional tangent space of $\SE$. It removes the $2K{-}3 = 5$ non-rigid components, which carries $5/8$ of the error energy. Slot errors from a shared backbone are correlated, so this is a motivation, not a guarantee. A second consideration is that predicting the visible corners in image coordinates preserves the spatial correspondence between output and input, whereas a rotation token $\Delta\theta_t$ turns an angular error into a position error scaled by the object's moment arm $\|\tilde{\mathbf{k}}_{0,i}\|_2$. Section~\ref{sec:results} tests this choice directly.

\subsection{Pre-L1 versus E2E-L1 Supervision}
\label{sec:method_jacobian}
All trajectory decoders are trained with
\begin{align}
\label{eq:unified_loss}
\mathcal{L}(\alpha) &= \mathcal{L}_{\text{CE}}^{\text{ls}=0.02}(\mathbf{L}, \mathbf{z}^*) + 0.5\,\mathcal{L}_{\text{CE}}(\mathbf{m}, m^*) \nonumber\\
&\quad + 15\,\alpha\,\mathcal{L}_{\text{SL1}}\bigl(\hat{\mathbf{Y}}, \mathbf{Y}^*\bigr) \nonumber\\
&\quad + 15\,(1-\alpha)\,\mathcal{L}_{\text{SL1}}\bigl(\Pi_{\text{Kabsch}}(\hat{\mathbf{Y}}, \mathbf{K}_0), \mathbf{Y}^*\bigr),
\end{align}
where the first term is token cross-entropy with label smoothing $0.02$, $\mathbf{m}$ are the logits of an auxiliary contact-mode head, and $\mathcal{L}_{\text{SL1}}$ is the Smooth-$L_1$ loss on coordinates divided by $512$. The CE head uses the first two terms only, without label smoothing: Eq.~\eqref{eq:vflash_ce} plus the same auxiliary contact-mode term. With $\alpha{=}1$ (\textbf{Pre-L1}) the Smooth-$L_1$ loss acts on the raw soft-token outputs and the projection is applied only at test time; Pre-L1 is therefore the Soft-L1 head followed by $\Pi_{\text{Kabsch}}$. With $\alpha{=}0$ (\textbf{E2E-L1}) the loss acts on the projected trajectory, and its gradient flows through $\Pi_{\text{Kabsch}}$. Both variants have $2{,}050{,}733$ parameters and the same inference cost.

\begin{proposition}[Rank of the projection Jacobian]
\label{prop:jacobian_nullspace}
For $K{=}4$ planar keypoints, $\hat{\mathbf{y}}_t \in \mathbb{R}^8$, the image of $\Pi_{\text{Kabsch}}(\cdot, \mathbf{K}_0)$ is the 3-dimensional manifold $\mathcal{M}(\mathbf{K}_0) \subset \mathbb{R}^8$ of rigid motions of $\mathbf{K}_0$. At any $\hat{\mathbf{y}}_t$ with $\alpha_t^2 + \beta_t^2 > 0$, the Jacobian $\mathbf{J}_{\Pi} = \partial \Pi_{\text{Kabsch}}(\hat{\mathbf{y}}_t, \mathbf{K}_0) / \partial \hat{\mathbf{y}}_t \in \mathbb{R}^{8 \times 8}$ has rank $3$. Its 5-dimensional null space consists of the perturbations that leave $(\hat{\mathbf{c}}_t, \hat{\theta}_t)$ unchanged to first order.
\end{proposition}
\begin{proof}
$\Pi_{\text{Kabsch}}$ factors as $\psi \circ \phi$, where $\phi(\hat{\mathbf{y}}_t) = (\hat{c}_{t,x}, \hat{c}_{t,y}, \hat{\theta}_t) \in \mathbb{R}^2 \times S^1$ and $\psi(\mathbf{c}, \theta)_i = \mathbf{c} + \mathbf{R}(\theta)\tilde{\mathbf{k}}_{0,i}$. The map $\phi$ is a submersion where $\alpha_t^2 + \beta_t^2 > 0$: the centroid depends on all coordinates with full rank 2, and a rigid rotation of the centered prediction changes $\hat{\theta}_t$ at unit rate. The map $\psi$ is an immersion for any non-degenerate shape $\mathbf{K}_0$. Hence $\text{rank}(\mathbf{J}_{\Pi}) = \text{rank}(\mathrm{d}\psi\,\mathrm{d}\phi) = 3$ and $\dim\ker\mathbf{J}_{\Pi} = 8 - 3 = 5$.
\end{proof}

Proposition~\ref{prop:jacobian_nullspace} is consistent with the trade-off we observe between the two variants. With Pre-L1, the Smooth-$L_1$ gradient constrains all $8$ keypoint coordinates, so each slot is trained to track its own corner. With E2E-L1, the loss gives no gradient along the $5$ non-rigid directions, so the slots are only constrained through the pose they jointly imply. On the continuous human demonstrations of \texttt{lerobot/pusht} this lowers trajectory error ($\PubSoftKabMapAde \to \PubEteMapAde\,\text{px}$). On ManiSkill3, where the decision depends on small differences between candidate actions, it raises regret ($\MsPreRegret \to \MsEteRegret$; Section~\ref{sec:results}). The proposition does not by itself predict the sign of either effect.

\begin{table*}[t]
\centering
\caption{ManiSkill3 \texttt{PushT-v1} test split, world frame (512 families $\times$ 4 candidate actions $\times$ 4 physics realizations $= 8{,}192$ rollouts; 5 seeds; family-cluster bootstrap, $B{=}2000$, 95\% CIs). The CE, Soft-L1 and MDN heads and their Kabsch-projected and $\text{SE}(2)$-token variants come from one training batch, with each projected variant evaluated on its parent's checkpoints; the other models come from the original benchmark run. Pre-reg.: arm of the pre-registered benchmark; Expl.: exploratory control added afterwards. Latency: median over 200 calls for one decision (batch of 4 candidates, $S{=}16$ samples where applicable). The pre-registered diffusion arm is omitted here because its implementation was faulty (Section~\ref{sec:grid_scaling}); its numbers are in Table~\ref{tab:superseded} and the corrected model is in Table~\ref{tab:v4_grid_scaling}.}
\label{tab:maniskill_main}
\resizebox{\textwidth}{!}{%
\begin{tabular}{llccccccc}
\toprule
\textbf{Model} & \textbf{Status} & \textbf{Regret $\downarrow$} & \textbf{Top-1 \% $\uparrow$} & \textbf{MAP-ADE $\downarrow$ (px)} & \textbf{MinADE$_{16}$ $\downarrow$ (px)} & \textbf{RB-Err $\downarrow$ (px)} & \textbf{RB-Viol \% $\downarrow$} & \textbf{Latency (ms)} \\
\midrule
\textbf{Procrustes-FPD, Pre-L1} & Expl. & \textbf{0.040 \ci{[0.034,0.046]}} & \textbf{60.2 \ci{[56.9,63.6]}} & \textbf{20.0 \ci{[19.5,20.5]}} & $-$ & \textbf{0.00 \ci{[0.00,0.00]}} & 0.0\% & 1.07 \\
Soft-L1 head & Pre-reg. & 0.042 \ci{[0.036,0.049]} & 59.1 \ci{[55.9,62.6]} & 20.9 \ci{[20.4,21.5]} & $-$ & 9.14 \ci{[8.85,9.42]} & 100.0\% & 1.07 \\
Procrustes-FPD, E2E-L1 & Expl. & 0.048 \ci{[0.041,0.054]} & 57.8 \ci{[54.5,61.1]} & 20.7 \ci{[20.1,21.3]} & $-$ & \textbf{0.00 \ci{[0.00,0.00]}} & 0.0\% & 1.07 \\
SE(2) tokens, Soft-L1 & Expl. & 0.049 \ci{[0.043,0.056]} & 56.0 \ci{[52.8,59.4]} & 22.9 \ci{[22.3,23.4]} & $-$ & \textbf{0.00 \ci{[0.00,0.00]}} & 0.0\% & 1.05 \\
\midrule
CE head (VFlash-style) & Pre-reg. & 0.042 \ci{[0.037,0.049]} & 57.8 \ci{[54.5,61.2]} & 25.6 \ci{[24.9,26.3]} & 25.2 \ci{[24.5,25.8]} & 14.40 \ci{[14.00,14.78]} & 100.0\% & 1.21 \\
CE head + Kabsch & Expl. & 0.044 \ci{[0.037,0.051]} & 57.5 \ci{[54.1,60.9]} & 23.2 \ci{[22.6,23.9]} & 21.5 \ci{[21.0,22.2]} & \textbf{0.00 \ci{[0.00,0.00]}} & 0.0\% & 1.21 \\
SE(2) tokens, CE & Expl. & 0.049 \ci{[0.042,0.055]} & 53.8 \ci{[50.5,57.0]} & 25.5 \ci{[24.9,26.2]} & 24.0 \ci{[23.4,24.5]} & \textbf{0.00 \ci{[0.00,0.00]}} & 0.0\% & 1.40 \\
\midrule
MDN head (4 branches) & Pre-reg. & 0.044 \ci{[0.038,0.051]} & 57.0 \ci{[53.6,60.4]} & 25.0 \ci{[24.4,25.7]} & 19.8 \ci{[19.3,20.4]} & 7.16 \ci{[6.93,7.39]} & 100.0\% & 1.18 \\
MDN head + Kabsch & Expl. & 0.046 \ci{[0.039,0.053]} & 56.3 \ci{[52.9,59.6]} & 24.6 \ci{[23.9,25.2]} & 19.2 \ci{[18.6,19.7]} & \textbf{0.00 \ci{[0.00,0.00]}} & 0.0\% & 1.18 \\
SE(2) tokens, MDN & Expl. & 0.041 \ci{[0.036,0.047]} & 57.1 \ci{[53.5,60.7]} & 24.4 \ci{[23.8,25.0]} & \textbf{18.3 \ci{[17.8,18.8]}} & \textbf{0.00 \ci{[0.00,0.00]}} & 0.0\% & 1.18 \\
\midrule
Bi-FPD (mode-routed MDN) & Pre-reg. & 0.048 \ci{[0.041,0.056]} & 55.7 \ci{[52.4,59.1]} & 24.3 \ci{[23.6,25.0]} & 20.0 \ci{[19.6,20.5]} & 6.74 \ci{[6.55,6.94]} & 99.9\% & 1.52 \\
Direct utility regression & Pre-reg. & 0.052 \ci{[0.045,0.059]} & 52.0 \ci{[48.6,55.5]} & 34.7 \ci{[34.1,35.3]} & $-$ & 13.37 \ci{[13.13,13.60]} & 100.0\% & 0.81 \\
Contact-mode classifier & Pre-reg. & 0.060 \ci{[0.053,0.067]} & 48.9 \ci{[45.6,52.5]} & $-$ & $-$ & $-$ & $-$ & 1.25 \\
Autoregressive (KV cache) & Pre-reg. & 0.076 \ci{[0.068,0.085]} & 44.1 \ci{[41.3,47.1]} & 37.2 \ci{[36.4,38.1]} & 27.3 \ci{[26.7,28.0]} & 7.83 \ci{[7.66,8.00]} & 99.1\% & 11.98 \\
\bottomrule
\end{tabular}%
}
\end{table*}

\section{Experimental Setup}
\label{sec:experiments}

\subsection{Benchmarks}
\textbf{ManiSkill3 \texttt{PushT-v1}.} We build a counterfactual benchmark on ManiSkill3~\cite{maniskill3} (SAPIEN/PhysX on CPU, Panda arm with a stick tool). It has $512$ training, $256$ validation and $512$ test families. A family fixes the initial T-block and pusher state; from it, $4$ candidate actions are each rolled out under $4$ hidden physics realizations that vary friction ($[0.05, 2.5]$), center-of-gravity offset, mass and inertia. This gives $16$ rollouts per family and $8{,}192$ test rollouts. The test split was generated after the benchmark was registered and was not used for any model selection. As a reference, an oracle that knows the true distribution over four contact modes and scores each action by its mean training utility per mode reaches $0.069\ci{[0.059,\,0.080]}$ regret ($47.1\%$ Top-1). Ranking actions therefore requires predicting trajectory geometry, not only the contact mode.

\textbf{\texttt{lerobot/pusht}.} The second benchmark uses the $206$ human-teleoperated demonstrations of the simulated Push-T task released with Diffusion Policy~\cite{chi2023diffusion} and distributed through LeRobot~\cite{lerobot} ($25{,}650$ frames at $10\,\text{Hz}$). Splits are episode-disjoint: episodes $0$--$139$ for training ($7{,}663$ windows), $140$--$165$ for validation ($1{,}487$ windows) and $166$--$205$ for testing ($2{,}384$ windows). The same test split was used in every stage of the study. The four T-block keypoints (junction, stem tip, left and right bar ends) are midpoints of the eight vertices given in \texttt{lerobot/pusht\_keypoints}; their pairwise distances vary by less than $10^{-5}\,\text{px}$ over all frames. Utility is computed against the fixed goal pose $[256, 256, \pi/4]$ of the environment. Because every window has only one recorded outcome, this benchmark measures trajectory error, not decision regret.

\subsection{Metrics}
\label{sec:metrics}
Coordinates are in workspace units: the planar workspace is mapped to $[0, 512]$, and we report distances in these units as px. The utility of a predicted outcome is
\begin{equation}
U = \text{clip}\!\left(1 - \frac{d_{\text{final}}}{\max(d_{\text{init}}, 30)},\, 0,\, 1\right)\exp\!\left(-\Bigl(\frac{\Delta\theta}{0.35}\Bigr)^{2}\right),
\end{equation}
where $d$ is the mean keypoint distance to the goal and $\Delta\theta$ is the wrapped orientation error of the junction-to-stem-tip vector.
\begin{itemize}
\item \textbf{Regret}: for each test family, the true mean utility (over the four realizations) of the best action minus that of the action the model selects, averaged over families. \textbf{Top-1}: the percentage of families in which the selected action is the best one.
\item \textbf{MAP-ADE}: the mean Euclidean error over the $10$ steps and $5$ points (four keypoints and the pusher) of the model's point prediction, averaged over realizations. \textbf{MinADE$_{16}$}: the same for the best of $16$ samples (stochastic decoders only).
\item \textbf{RB-Err}: the mean absolute deviation of the six pairwise keypoint distances from those of the true shape. \textbf{RB-Viol}: the percentage of predicted trajectories whose per-step deviation exceeds $2.5\,\text{px}$ at any step.
\item \textbf{Brier}: the Brier score of the predicted contact-mode distribution. \textbf{U-MAE}: the absolute error of the predicted utility (\texttt{lerobot/pusht}).
\end{itemize}

\subsection{Models}
All models share the same context encoder: a small convolutional image encoder and a 3-layer transformer of width $d{=}192$ over eight prefix tokens (image features, keypoint history, relative geometry, goal and action). They differ only in the output head and decoding procedure. The CE head (VFlash-style) is trained with Eq.~\eqref{eq:vflash_ce}. The Soft-L1 head adds the soft-token Smooth-$L_1$ term ($\alpha{=}1$), and Procrustes-FPD adds the Kabsch projection (Pre-L1 or E2E-L1). The MDN head predicts a 4-component mixture of trajectories, and Bi-FPD routes its components by predicted contact mode. The $\SE$-token variants of the CE, Soft-L1 and MDN heads predict the block pose $(\Delta x_t, \Delta y_t, \Delta\theta_t)$ and the pusher with 50 slots instead of 100 keypoint slots. The iterative baselines are an autoregressive decoder that emits one time step (all $10$ coordinates) per step, with a key-value cache ($10$ sequential steps), and a DDIM diffusion model over whole trajectories ($10$ steps unless stated). Both draw $16$ samples per candidate. We also include a direct utility regressor and a contact-mode classifier that score actions without predicting a trajectory. Parameter counts range from $1.9$M to $4.2$M, mainly because of the output layers. The slot heads share one 1001-way output layer across slots (about $0.19$M parameters), whereas the autoregressive and mixture heads predict all $10$ coordinates of a step from one token ($d \times 10 \times 1001 \approx 1.9$M parameters per output layer). Latency is the median of $200$ calls for one decision, i.e.\ a batch of four candidate actions with $S{=}16$ samples where applicable, on one GPU. Latencies in Tables~\ref{tab:maniskill_main} and~\ref{tab:v4_grid_scaling} come from separate measurement runs (the latter includes the frame transform) and should be compared only within a table. Table~\ref{tab:names} maps model names to identifiers in the released code.

\subsection{Study Design and Statistics}
\begin{enumerate}
\item \textbf{Pre-registered benchmark.} Twelve models, including the CE, Soft-L1, MDN, Bi-FPD, autoregressive, diffusion, direct-regression and classifier models above, were registered with their hypotheses before the ManiSkill3 test split was generated. Learning rate, weight decay, dropout and loss weights were selected on the validation split only and then frozen. The remaining four registered models, a CE head with an in-block contact-mode token and three Bi-FPD ablations, address a question outside this paper; we omit them from the tables, and their results will be included in the released files.
\item \textbf{Exploratory controls.} The Kabsch-projected variants, the $\SE$-token variants, E2E-L1 and all object-frame models were added afterwards. They reuse the frozen hyperparameters of their parent models without re-tuning, and we label them as exploratory.
\item \textbf{Matched checkpoints.} The CE, Soft-L1 and MDN heads were retrained together with their projected and $\SE$-token variants, and each projected variant is evaluated on exactly the same checkpoints as its unprojected parent. Retraining the CE head with identical seeds ($0$--$4$) and configuration moved its mean test regret from $0.0457$ to $0.0423$, a difference of $0.0034$ caused by non-deterministic CUDA kernels. The Soft-L1 head moved from $0.0424$ to $0.0423$, and the MDN head from $0.0440$ to $0.0443$. Family-cluster bootstrap intervals condition on the trained checkpoints, so regret differences below about $0.0035$ in Tables~\ref{tab:maniskill_main}--\ref{tab:se2_canonical} are within run-to-run variation. Bi-FPD, the autoregressive model, the direct regressor and the classifier were not retrained and come from the original benchmark run. Their regret gaps to Pre-L1 ($0.0082$ and larger) exceed this retraining shift.
\item \textbf{Registered grid.} After the exploratory object-frame results, we registered a $2{\times}5$ frame $\times$ decoder grid with hypotheses H1--H5, a corrected diffusion model and a data-scaling sweep, and ran it on the same test families. The grid uses a single training path for all decoders, removes the RGB input in both frames, and uses a seed $\times$ family hierarchical bootstrap that also resamples training seeds.
\item \textbf{Multiple comparisons.} Each table of paired tests applies one Holm step-down correction across all competitors it reports against its reference, including weak and close ones, plus the superseded diffusion arm; family sizes are stated in each table. All bootstraps use $B{=}2000$ resamples. A reported $p{=}0.0005$ means that no resample crossed zero.
\end{enumerate}

\begin{table*}[t]
\centering
\caption{\texttt{lerobot/pusht} test split, world frame (40 test episodes, $2{,}384$ windows; 3 seeds; episode-cluster bootstrap, $B{=}2000$, 95\% CIs; fixed goal pose $[256, 256, \pi/4]$). Raw: unprojected decoder output; Kabsch: after projection onto the rigid-body manifold. For the Soft-L1 head the Kabsch column is Procrustes-FPD (Pre-L1). Errors in pixels. The superseded diffusion arm is reported in Table~\ref{tab:superseded}.}
\label{tab:public_pusht}
\resizebox{\textwidth}{!}{%
\begin{tabular}{lcccccccc}
\toprule
\textbf{Model} & \textbf{MAP-ADE Raw $\downarrow$} & \textbf{MAP-ADE Kabsch $\downarrow$} & \textbf{MinADE$_{16}$ Raw $\downarrow$} & \textbf{MinADE$_{16}$ Kabsch $\downarrow$} & \textbf{RB-Err Raw $\downarrow$} & \textbf{RB-Err Kabsch $\downarrow$} & \textbf{U-MAE Kabsch $\downarrow$} \\
\midrule
\textbf{Procrustes-FPD, E2E-L1} & \textbf{10.17 \ci{[9.33,11.10]}} & \textbf{10.17 \ci{[9.33,11.10]}} & $-$ & $-$ & \textbf{0.00 \ci{[0.00,0.00]}} & \textbf{0.000 \ci{[0.000,0.000]}} & 0.068 \ci{[0.061,0.076]} \\
Soft-L1 head & 12.78 \ci{[11.90,13.76]} & 11.76 \ci{[10.88,12.73]} & $-$ & $-$ & 6.50 \ci{[6.06,6.92]} & \textbf{0.000 \ci{[0.000,0.000]}} & 0.086 \ci{[0.077,0.095]} \\
SE(2) tokens, Soft-L1 & 13.14 \ci{[12.27,14.16]} & 13.14 \ci{[12.27,14.16]} & $-$ & $-$ & \textbf{0.00 \ci{[0.00,0.00]}} & \textbf{0.000 \ci{[0.000,0.000]}} & 0.071 \ci{[0.063,0.080]} \\
SE(2) tokens, MDN & 13.92 \ci{[12.81,15.13]} & 13.92 \ci{[12.81,15.13]} & \textbf{12.59 \ci{[11.53,13.77]}} & \textbf{12.59 \ci{[11.53,13.77]}} & \textbf{0.00 \ci{[0.00,0.00]}} & \textbf{0.000 \ci{[0.000,0.000]}} & 0.064 \ci{[0.056,0.073]} \\
CE head (VFlash-style) & 16.53 \ci{[15.46,17.68]} & 14.38 \ci{[13.43,15.40]} & 17.97 \ci{[16.75,19.27]} & 14.94 \ci{[13.83,16.11]} & 9.97 \ci{[9.25,10.69]} & \textbf{0.000 \ci{[0.000,0.000]}} & 0.067 \ci{[0.060,0.075]} \\
MDN head (4 branches) & 16.35 \ci{[15.06,17.86]} & 15.61 \ci{[14.31,17.11]} & 15.04 \ci{[13.77,16.53]} & 14.21 \ci{[12.94,15.68]} & 6.33 \ci{[5.71,7.03]} & \textbf{0.000 \ci{[0.000,0.000]}} & 0.080 \ci{[0.070,0.090]} \\
Direct utility regression & 16.40 \ci{[15.68,17.13]} & 15.11 \ci{[14.43,15.81]} & $-$ & $-$ & 8.74 \ci{[8.27,9.29]} & \textbf{0.000 \ci{[0.000,0.000]}} & 0.080 \ci{[0.072,0.087]} \\
Bi-FPD (mode-routed MDN) & 16.75 \ci{[15.44,18.25]} & 15.79 \ci{[14.55,17.22]} & 15.19 \ci{[14.05,16.49]} & 14.07 \ci{[12.99,15.32]} & 6.09 \ci{[5.44,6.77]} & \textbf{0.000 \ci{[0.000,0.000]}} & \textbf{0.059 \ci{[0.051,0.067]}} \\
Autoregressive (KV cache) & 25.43 \ci{[23.31,27.79]} & 24.11 \ci{[22.07,26.42]} & 24.68 \ci{[22.84,26.73]} & 21.34 \ci{[19.62,23.26]} & 6.84 \ci{[6.30,7.45]} & \textbf{0.000 \ci{[0.000,0.000]}} & 0.069 \ci{[0.060,0.079]} \\
\bottomrule
\end{tabular}%
}
\end{table*}

\begin{figure*}[t]
\centering
\includegraphics[width=0.96\textwidth]{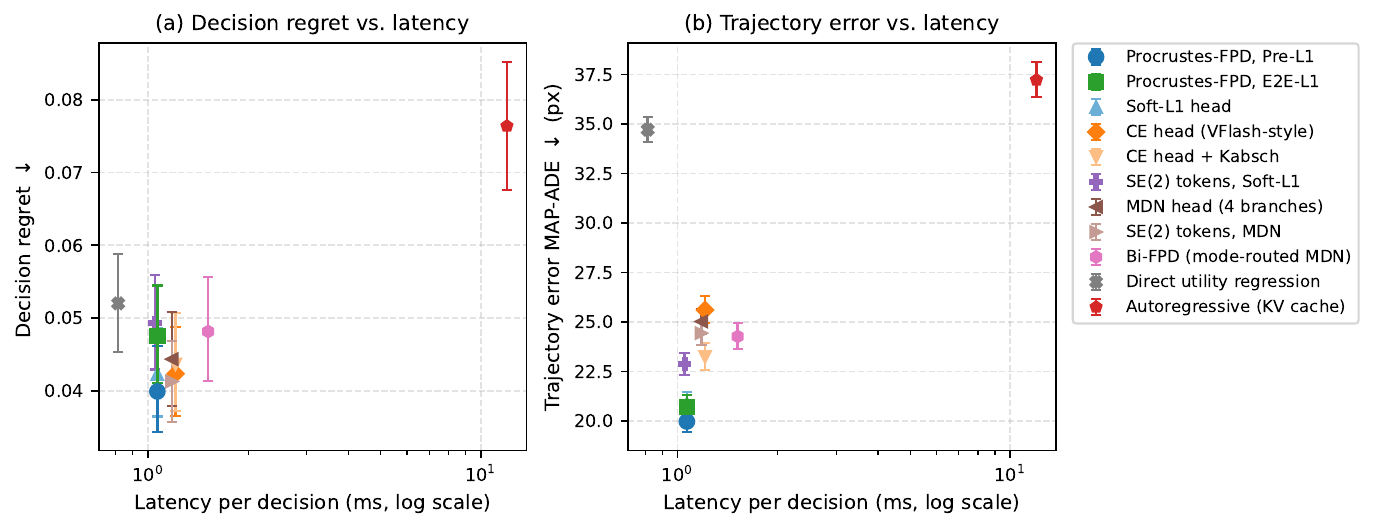}
\caption{ManiSkill3 test split, world frame (512 families, 5 seeds, 95\% family-cluster bootstrap CIs). \textbf{(a)} Decision regret and \textbf{(b)} trajectory error (MAP-ADE) against latency per decision (log scale).}
\label{fig:pareto_maniskill}
\end{figure*}

\begin{figure*}[t]
\centering
\includegraphics[width=0.92\textwidth]{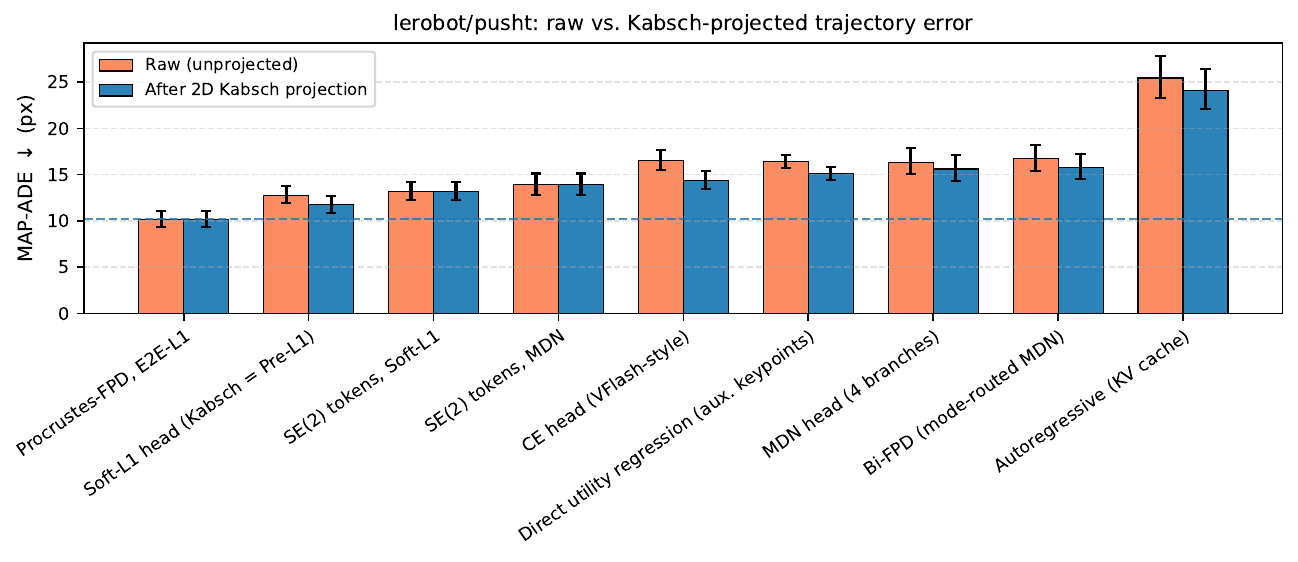}
\caption{\texttt{lerobot/pusht} test split (40 episodes, $2{,}384$ windows, 3 seeds, 95\% episode-cluster bootstrap CIs): trajectory error of the raw decoder output (orange) and after 2D Kabsch projection (blue). The dashed line marks Procrustes-FPD (E2E-L1).}
\label{fig:public_bars}
\end{figure*}

\section{Results}
\label{sec:results}

\subsection{ManiSkill3, World Frame}
Table~\ref{tab:maniskill_main}, the top half of Table~\ref{tab:paired_tests} and Fig.~\ref{fig:pareto_maniskill} report the matched world-frame comparison.

\textbf{Trajectory prediction beats direct scoring.} Procrustes-FPD (Pre-L1) has lower regret (\MsPreRegret{}, $1.07\,\text{ms}$) than the KV-cached autoregressive model (\MsArRegret{}, $11.98\,\text{ms}$), the direct utility regressor ($0.052$) and the contact-mode classifier ($0.060$; all $\pholm{14}{=}0.0070$).

\textbf{Soft-token expectation improves trajectory error.} The Soft-L1 head lowers MAP-ADE from \MsPureCeMapAde\,px (CE head) to \MsRawSoftMapAde\,px, with non-overlapping 95\% CIs; regret is unchanged (\MsRawSoftRegret{} vs.\ \MsPureCeRegret{}). Projecting the Soft-L1 output with $\Pi_{\text{Kabsch}}$ (Pre-L1) removes all rigid-body error (from \MsRawSoftRbErr\,px and $100\%$ violations to zero) and lowers MAP-ADE by a further $0.91\,\text{px}$ to \MsPreMapAde\,px ($\pholm{13}{=}0.0065$). Pre-L1 has the lowest MAP-ADE of all models, significantly below every other model in Table~\ref{tab:maniskill_main} ($\pholm{13}{=}0.0065$), e.g.\ $5.63\,\text{px}$ below the CE head and $3.26\,\text{px}$ below the CE head with the same projection.

\textbf{The best decision regrets are statistically tied.} Pre-L1 has significantly lower regret than the $\SE$-token Soft-L1 head ($\Delta{=}{+}0.0094$), the $\SE$-token CE head ($+0.0087$), Bi-FPD ($+0.0082$) and E2E-L1 ($+0.0076$; all $\pholm{14}{=}0.0070$). Its advantage over the CE head ($+0.0024$, $\pholm{14}{=}0.516$), the $\SE$-token MDN head ($+0.0014$, $\pholm{14}{=}0.531$), the unprojected Soft-L1 head ($+0.0024$, unadjusted $p{=}0.025$, $\pholm{14}{=}0.125$) and the MDN head ($+0.0044$, $p{=}0.028$, $\pholm{14}{=}0.125$) is not significant after correction. Most of these differences are also within the run-to-run variation of about $0.0035$. In the world frame, therefore, the decoder head affects trajectory accuracy and rigidity much more than it affects decisions.

\begin{figure}[t]
\centering
\includegraphics[width=\columnwidth]{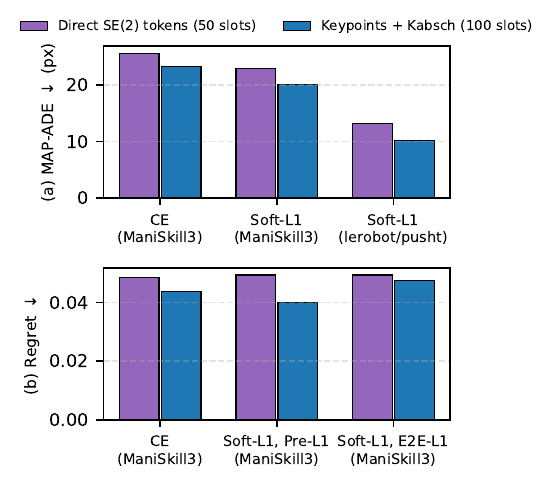}
\caption{Direct $\SE$ pose tokens $(\Delta x_t, \Delta y_t, \Delta\theta_t)$ with 50 slots (purple) against 100 keypoint slots with Kabsch projection (blue), with identical backbones and losses, world frame. \textbf{(a)} MAP-ADE. \textbf{(b)} Regret on ManiSkill3.}
\label{fig:se2_ablation}
\end{figure}

\textbf{Keypoints with Procrustes fitting beat direct pose tokens.} With identical backbones and losses, $100$ keypoint slots followed by $\Pi_{\text{Kabsch}}$ outperform $50$ direct $\SE$ slots in MAP-ADE ($\MsPreMapAde$ vs.\ $\MsSeTwoSoftMapAde\,\text{px}$, $\pholm{13}{=}0.0065$) and in regret ($\MsPreRegret$ vs.\ $\MsSeTwoSoftRegret$, $\pholm{14}{=}0.0070$), consistent with findings in 6-DoF pose estimation~\cite{peng2019pvnet,zhou2019continuity} (Fig.~\ref{fig:se2_ablation}).

\begin{table*}[t]
\centering
\caption{Paired cluster-bootstrap tests ($B{=}2000$; two-sided $p$, with Holm-corrected $p_{\text{Holm}}$ in parentheses). $\Delta$ is competitor minus reference, so $\Delta > 0$ favors the reference. Top: ManiSkill3, world frame, against Procrustes-FPD (Pre-L1), on regret and MAP-ADE. Bottom: \texttt{lerobot/pusht}, world frame, against Procrustes-FPD (E2E-L1), on raw and Kabsch-projected MAP-ADE. Each Holm family contains all competitors shown plus the superseded diffusion arm (not shown; Table~\ref{tab:superseded}): 14 (regret) and 13 (MAP-ADE) comparisons on ManiSkill3, 9 on \texttt{lerobot/pusht}. Removing the superseded arm could only lower the other adjusted $p$-values.}
\label{tab:paired_tests}
\resizebox{\textwidth}{!}{%
\begin{tabular}{llcccccc}
\toprule
\textbf{Benchmark} & \textbf{Competitor} & \textbf{Regret / Raw $\Delta$} & \textbf{95\% CI} & \textbf{$p$ ($p_{\text{Holm}}$)} & \textbf{MAP-ADE / Kabsch $\Delta$ (px)} & \textbf{95\% CI} & \textbf{$p$ ($p_{\text{Holm}}$)} \\
\midrule
ManiSkill3 (vs.\ Pre-L1) & Soft-L1 head & $+0.0024$ & $[+0.0003,\,+0.0047]$ & $0.0250\;(0.1250)$ & $+0.91$ & $[+0.86,\,+0.96]$ & $0.0005\;(0.0065)$ \\
ManiSkill3 (vs.\ Pre-L1) & Procrustes-FPD, E2E-L1 & $+0.0076$ & $[+0.0040,\,+0.0114]$ & $0.0005\;(0.0070)$ & $+0.74$ & $[+0.51,\,+0.98]$ & $0.0005\;(0.0065)$ \\
ManiSkill3 (vs.\ Pre-L1) & SE(2) tokens, Soft-L1 & $+0.0094$ & $[+0.0054,\,+0.0138]$ & $0.0010\;(0.0070)$ & $+2.91$ & $[+2.65,\,+3.16]$ & $0.0005\;(0.0065)$ \\
ManiSkill3 (vs.\ Pre-L1) & CE head (VFlash-style) & $+0.0024$ & $[-0.0020,\,+0.0064]$ & $0.2580\;(0.5160)$ & $+5.63$ & $[+5.31,\,+5.97]$ & $0.0005\;(0.0065)$ \\
ManiSkill3 (vs.\ Pre-L1) & CE head + Kabsch & $+0.0037$ & $[+0.0000,\,+0.0073]$ & $0.0500\;(0.1500)$ & $+3.26$ & $[+2.97,\,+3.57]$ & $0.0005\;(0.0065)$ \\
ManiSkill3 (vs.\ Pre-L1) & SE(2) tokens, CE & $+0.0087$ & $[+0.0043,\,+0.0130]$ & $0.0005\;(0.0070)$ & $+5.55$ & $[+5.23,\,+5.88]$ & $0.0005\;(0.0065)$ \\
ManiSkill3 (vs.\ Pre-L1) & MDN head (4 branches) & $+0.0044$ & $[+0.0005,\,+0.0087]$ & $0.0280\;(0.1250)$ & $+5.05$ & $[+4.72,\,+5.37]$ & $0.0005\;(0.0065)$ \\
ManiSkill3 (vs.\ Pre-L1) & MDN head + Kabsch & $+0.0061$ & $[+0.0021,\,+0.0103]$ & $0.0030\;(0.0180)$ & $+4.59$ & $[+4.25,\,+4.92]$ & $0.0005\;(0.0065)$ \\
ManiSkill3 (vs.\ Pre-L1) & SE(2) tokens, MDN & $+0.0014$ & $[-0.0031,\,+0.0060]$ & $0.5310\;(0.5310)$ & $+4.46$ & $[+4.09,\,+4.87]$ & $0.0005\;(0.0065)$ \\
ManiSkill3 (vs.\ Pre-L1) & Bi-FPD (mode-routed MDN) & $+0.0082$ & $[+0.0036,\,+0.0131]$ & $0.0005\;(0.0070)$ & $+4.29$ & $[+3.96,\,+4.67]$ & $0.0005\;(0.0065)$ \\
ManiSkill3 (vs.\ Pre-L1) & Direct utility regression & $+0.0121$ & $[+0.0056,\,+0.0187]$ & $0.0005\;(0.0070)$ & $+14.73$ & $[+14.32,\,+15.15]$ & $0.0005\;(0.0065)$ \\
ManiSkill3 (vs.\ Pre-L1) & Contact-mode classifier & $+0.0198$ & $[+0.0125,\,+0.0273]$ & $0.0005\;(0.0070)$ & $-$ & $-$ & $-$ \\
ManiSkill3 (vs.\ Pre-L1) & Autoregressive (KV cache) & $+0.0365$ & $[+0.0293,\,+0.0440]$ & $0.0005\;(0.0070)$ & $+17.24$ & $[+16.62,\,+17.94]$ & $0.0005\;(0.0065)$ \\
\midrule
\texttt{lerobot/pusht} (vs.\ E2E-L1) & Soft-L1 head & $+2.61\,\text{px}$ & $[+2.37,\,+2.87]$ & $0.0005\;(0.0045)$ & $+1.59\,\text{px}$ & $[+1.35,\,+1.83]$ & $0.0005\;(0.0045)$ \\
\texttt{lerobot/pusht} (vs.\ E2E-L1) & SE(2) tokens, Soft-L1 & $+2.97\,\text{px}$ & $[+2.62,\,+3.32]$ & $0.0005\;(0.0045)$ & $+2.97\,\text{px}$ & $[+2.62,\,+3.32]$ & $0.0005\;(0.0045)$ \\
\texttt{lerobot/pusht} (vs.\ E2E-L1) & SE(2) tokens, MDN & $+3.74\,\text{px}$ & $[+3.18,\,+4.29]$ & $0.0005\;(0.0045)$ & $+3.74\,\text{px}$ & $[+3.18,\,+4.29]$ & $0.0005\;(0.0045)$ \\
\texttt{lerobot/pusht} (vs.\ E2E-L1) & CE head (VFlash-style) & $+6.36\,\text{px}$ & $[+5.81,\,+6.94]$ & $0.0005\;(0.0045)$ & $+4.21\,\text{px}$ & $[+3.77,\,+4.66]$ & $0.0005\;(0.0045)$ \\
\texttt{lerobot/pusht} (vs.\ E2E-L1) & MDN head (4 branches) & $+6.18\,\text{px}$ & $[+5.53,\,+6.87]$ & $0.0005\;(0.0045)$ & $+5.44\,\text{px}$ & $[+4.81,\,+6.10]$ & $0.0005\;(0.0045)$ \\
\texttt{lerobot/pusht} (vs.\ E2E-L1) & Direct utility regression & $+6.23\,\text{px}$ & $[+5.75,\,+6.69]$ & $0.0005\;(0.0045)$ & $+4.94\,\text{px}$ & $[+4.45,\,+5.42]$ & $0.0005\;(0.0045)$ \\
\texttt{lerobot/pusht} (vs.\ E2E-L1) & Bi-FPD (mode-routed MDN) & $+6.58\,\text{px}$ & $[+5.87,\,+7.33]$ & $0.0005\;(0.0045)$ & $+5.62\,\text{px}$ & $[+4.96,\,+6.34]$ & $0.0005\;(0.0045)$ \\
\texttt{lerobot/pusht} (vs.\ E2E-L1) & Autoregressive (KV cache) & $+15.26\,\text{px}$ & $[+13.80,\,+16.87]$ & $0.0005\;(0.0045)$ & $+13.94\,\text{px}$ & $[+12.55,\,+15.50]$ & $0.0005\;(0.0045)$ \\
\bottomrule
\end{tabular}%
}
\end{table*}

\subsection{\texttt{lerobot/pusht}, World Frame}
Table~\ref{tab:public_pusht}, the bottom half of Table~\ref{tab:paired_tests} and Fig.~\ref{fig:public_bars} report the 3-seed evaluation on the $40$ test episodes.

\textbf{Projection helps every Cartesian decoder.} Applying $\Pi_{\text{Kabsch}}$ at test time lowers MAP-ADE for the CE head ($\PubPureCeRawMapAde \to \PubPureCeKabMapAde\,\text{px}$), the Soft-L1 head ($\PubSoftRawMapAde \to \PubSoftKabMapAde\,\text{px}$), the MDN head ($16.35 \to 15.61\,\text{px}$), Bi-FPD ($\PubBiFpdRawMapAde \to \PubBiFpdKabMapAde\,\text{px}$), the direct regressor's auxiliary keypoints ($16.40 \to 15.11\,\text{px}$) and the autoregressive model ($\PubArRawMapAde \to \PubArKabMapAde\,\text{px}$). As on ManiSkill3, the Soft-L1 head improves on the CE head ($\PubPureCeRawMapAde$ vs.\ $\PubSoftRawMapAde\,\text{px}$ raw, non-overlapping CIs).

\textbf{Training through the projection gives the lowest trajectory error here.} E2E-L1 reaches \PubEteMapAde\,px, lower than every other model in Table~\ref{tab:public_pusht} ($\pholm{9}{=}0.0045$), including Pre-L1 (\PubSoftKabMapAde\,px, $\Delta{=}{+}1.59\,\text{px}$) and the $\SE$-token Soft-L1 head (\PubSeTwoSoftMapAde\,px, $\Delta{=}{+}2.97\,\text{px}$). On utility error, the mode-routed Bi-FPD is best ($0.059$ U-MAE vs.\ $0.068$ for E2E-L1). Together with the ManiSkill3 results, this shows that Pre-L1 and E2E-L1 trade places: E2E-L1 gives lower trajectory error on human demonstrations, while Pre-L1 gives lower regret on the counterfactual benchmark. Both were exploratory additions, so we report them as an ablation and do not select a single winner.

\subsection{The Object Frame}
\label{sec:canon}
In Table~\ref{tab:maniskill_main}, all eleven single-pass world-frame models have regret between $0.040$ and $0.049$. One possible reason is a noise floor: the regret targets are means over only four realizations, $\bar{u}_{f,a} = \frac{1}{4}\sum_{r=1}^4 u_{f,a,r}$. Bootstrapping the four realizations over the $512$ test families, however, puts this floor at only $0.0047$ regret (std $0.0009$). A second possibility is overfitting, and a world-frame Soft-L1 model (seed $0$) shows a large generalization gap:
\begin{itemize}
\item \textbf{Training split} ($512$ families): regret $0.0064$, Top-1 $82.6\%$, MAP-ADE $7.65\,\text{px}$.
\item \textbf{Validation split} ($256$ families): regret $0.0380$, Top-1 $58.2\%$, MAP-ADE $23.72\,\text{px}$.
\item \textbf{Test split} ($512$ families): regret $0.0484$, Top-1 $56.1\%$, MAP-ADE $22.66\,\text{px}$.
\end{itemize}
Planar pushing is equivariant to rigid motions of the scene, but a model that sees raw world coordinates from $512$ families can fit the absolute initial poses $(\mathbf{c}_0, \theta_0)$ instead of the pose-independent dynamics. We hypothesized that this explains the gap.

\begin{figure*}[t]
\centering
\includegraphics[width=0.94\textwidth]{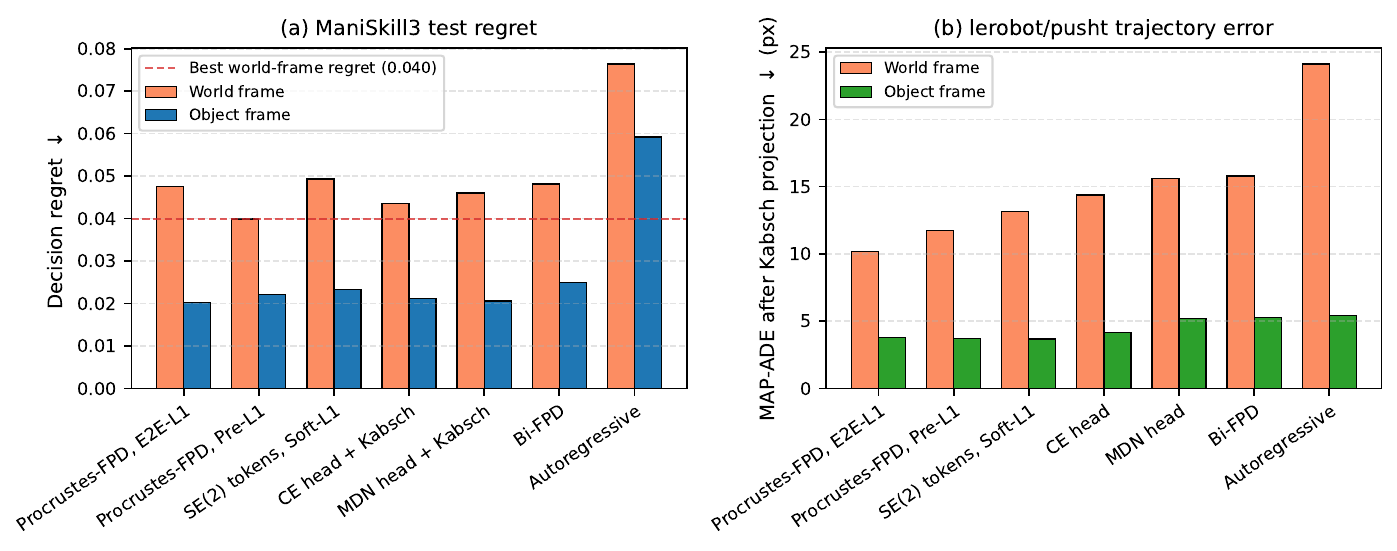}
\caption{World frame (orange) against object frame (blue/green) for each decoder. \textbf{(a)} ManiSkill3 test regret (512 families, 5 seeds); the dashed line marks the best world-frame regret. \textbf{(b)} \texttt{lerobot/pusht} MAP-ADE after Kabsch projection (40 episodes, 3 seeds); the object frame lowers it by $63\%$--$78\%$.}
\label{fig:se2_canonical}
\end{figure*}

To test this, we wrap each decoder in a fixed change of frame. Let $\mathbf{c}_0$ be the centroid of the observed initial keypoints and $\mathbf{R}_0$ the orientation of the stem-to-bar axis. All 2D inputs $\mathbf{p}$ (keypoint history, goal keypoints, pusher positions and action endpoints) and all targets are mapped into the object frame before tokenization, and predictions are mapped back:
\begin{align}
\mathbf{p}_{\text{OF}} &= \mathbf{o} + s\,\mathbf{R}_0^{\top}(\mathbf{p} - \mathbf{c}_0), \nonumber\\
\mathbf{p}_{\text{world}} &= \mathbf{c}_0 + s^{-1}\,\mathbf{R}_0(\hat{\mathbf{p}}_{\text{OF}} - \mathbf{o}),
\label{eq:body_frame_canon}
\end{align}
with $\mathbf{o} = (256, 256)^{\top}$ and $s = 0.75$. With this scale, every object-frame coordinate in the dataset lies in $[30, 486]$, so no clipping occurs. Unless stated otherwise, object-frame models receive no RGB input. Table~\ref{tab:se2_canonical} and Fig.~\ref{fig:se2_canonical} report the results.

\textbf{The object frame halves regret for every single-pass decoder.} On ManiSkill3, E2E-L1 drops from $\MsEteRegret$ to $\mathbf{\BfMsEteRegret}\ci{\BfMsEteRegretCI}$ regret (Top-1 $57.8\% \to \BfMsEteTopOne\%$, MAP-ADE $\MsEteMapAde \to \BfMsEteMapAde\,\text{px}$), Pre-L1 from $\MsPreRegret$ to $\BfMsPreRegret\ci{\BfMsPreRegretCI}$ (MAP-ADE $\BfMsPreMapAde\,\text{px}$), the CE head with projection from $\MsPureCeKabRegret$ to $\BfMsPureCeKabRegret$ (MAP-ADE $\MsPureCeKabMapAde \to \BfMsPureCeKabMapAde\,\text{px}$), and the MDN head with projection from $0.046$ to $0.021$ (MAP-ADE $24.6 \to 12.92\,\text{px}$). On \texttt{lerobot/pusht}, the object frame lowers projected MAP-ADE by $63\%$--$78\%$ across decoders (Fig.~\ref{fig:se2_canonical}b).

\textbf{The gain comes from the frame, not from removing the image.} Since the object-frame models receive no RGB, we trained world-frame controls without RGB as well. They reach \WfNoRgbMsRegret{} regret and \WfNoRgbMsMapAde\,px MAP-ADE on ManiSkill3 (\MsPreRegret{} and \MsPreMapAde\,px with RGB) and \WfNoRgbPubKabMapAde\,px on \texttt{lerobot/pusht} (\PubSoftKabMapAde\,px with RGB). Removing the image in the world frame does not help. Conversely, feeding the unwarped world-camera image to the object-frame model raises regret from $\BfMsPreRegret$ to $0.029$ ($\Delta{=}{+}0.0066$, unadjusted $p{=}0.001$) and MAP-ADE from $\BfMsPreMapAde$ to $13.71\,\text{px}$ ($p{=}0.0005$). This is consistent with the image leaking the world pose back into the model. Rotating the image into the object frame with a differentiable warp removes most of this effect:
\begin{itemize}
\item On the top-down view of \texttt{lerobot/pusht}, E2E-L1 with the rotated image reaches $\mathbf{\BfPubOrthoEteMapAde\,\text{px}}$ projected MAP-ADE. This matches the object-frame model without image (\BfPubEteMapAde\,px, $p{=}0.495$) and is $6.45\,\text{px}$ lower than world-frame E2E-L1 (\PubEteMapAde\,px, $p{=}0.0005$).
\item On the oblique ($39^{\circ}$) camera of ManiSkill3, warping the image through the table-plane homography gives \BfRotRgbMsEteRegret{} regret and \BfRotRgbMsEteMapAde\,px MAP-ADE. This is better than world-frame E2E-L1 ($\MsEteRegret$, $p{=}0.0005$) but worse than the object-frame model without image (\BfMsEteRegret{}, $p{=}0.0005$). A plausible explanation, which we have not tested, is that perspective and lighting in the oblique view still reveal part of the world pose.
\end{itemize}
These $p$-values are unadjusted paired bootstrap tests of exploratory controls.

\textbf{Projection still gives the lowest trajectory error in the object frame.} On ManiSkill3, E2E-L1 ($11.35\,\text{px}$) has lower MAP-ADE than the projected CE head ($11.70\,\text{px}$), the projected MDN head ($12.92\,\text{px}$), projected Bi-FPD ($14.28\,\text{px}$) and the autoregressive model ($28.32\,\text{px}$; all $\pholm{16}{=}0.0080$), and is tied with Pre-L1 ($11.42\,\text{px}$, $\pholm{16}{=}0.062$). On \texttt{lerobot/pusht}, Pre-L1 ($3.71\,\text{px}$), E2E-L1 ($3.78\,\text{px}$) and $\SE$ tokens ($3.68\,\text{px}$) are statistically tied. Pre-L1 is lower than the CE head ($4.13\,\text{px}$), the MDN head ($5.19\,\text{px}$), Bi-FPD ($5.27\,\text{px}$) and the autoregressive model ($5.43\,\text{px}$; all $\pholm{10}{=}0.0050$). Once the world rotation $\theta_0$ is removed, $\SE$ tokens lose most of their disadvantage ($11.84\,\text{px}$ on ManiSkill3), so their moment-arm penalty matters mainly when rotations are large. On regret, the single-pass heads in the object frame lie between $0.020$ and $0.025$, and most pairwise differences are not significant.

\begin{table*}[t]
\centering
\caption{Object-frame (OF) canonicalization. Each model is trained and evaluated in the frame of the initial object pose $(\mathbf{c}_0, \mathbf{R}_0)$ and its predictions are mapped back to the world frame. Unless marked, OF models receive no RGB input; ``warped'', ``rotated'' and ``unwarped'' RGB variants are described in Section~\ref{sec:canon}. Gain: paired improvement over the same model in the world frame (Tables~\ref{tab:maniskill_main} and~\ref{tab:public_pusht}). Top: ManiSkill3 (5 seeds); $p_{\text{Holm}(16)}$ against OF Procrustes-FPD (E2E-L1). Bottom: \texttt{lerobot/pusht} (3 seeds); $p_{\text{Holm}(10)}$ against the OF Soft-L1 head after Kabsch projection (Pre-L1). Holm families include the superseded OF diffusion arm (Table~\ref{tab:superseded}), which is not shown.}
\label{tab:se2_canonical}
\resizebox{\textwidth}{!}{%
\begin{tabular}{lcccccccc}
\toprule
\textbf{ManiSkill3, OF (5 seeds)} & \textbf{Regret $\downarrow$} & \textbf{Top-1 \% $\uparrow$} & \textbf{MAP-ADE $\downarrow$ (px)} & \textbf{RB-Err $\downarrow$ (px)} & \textbf{Brier $\downarrow$} & \textbf{Gain $\Delta_{\text{WF}-\text{OF}}$ (Reg / ADE)} & \textbf{$p_{\text{Holm}(16)}$ (Reg / ADE)} \\
\midrule
\textbf{Procrustes-FPD, E2E-L1} & \textbf{0.020 \ci{[0.016,0.025]}} & \textbf{73.2 \ci{[69.8,76.4]}} & \textbf{11.35 \ci{[10.95,11.75]}} & \textbf{0.00 \ci{[0.00,0.00]}} & 0.188 \ci{[0.173,0.205]} & $+0.027$ / $+9.4\,\text{px}$ & \textbf{Reference} \\
Procrustes-FPD, Pre-L1 & 0.022 \ci{[0.018,0.027]} & 71.3 \ci{[68.0,74.6]} & 11.42 \ci{[11.02,11.81]} & \textbf{0.00 \ci{[0.00,0.00]}} & 0.202 \ci{[0.185,0.219]} & $+0.018$ / $+8.6\,\text{px}$ & $0.465$ / $0.0620$ \\
Soft-L1 head & 0.023 \ci{[0.019,0.028]} & 71.1 \ci{[67.9,74.5]} & 11.85 \ci{[11.46,12.24]} & 4.69 \ci{[4.57,4.81]} & 0.202 \ci{[0.185,0.219]} & $+0.019$ / $+9.0\,\text{px}$ & $0.045$ / $0.0080$ \\
SE(2) tokens, Soft-L1 & 0.023 \ci{[0.019,0.028]} & 72.5 \ci{[69.4,75.6]} & 11.84 \ci{[11.45,12.22]} & \textbf{0.00 \ci{[0.00,0.00]}} & 0.189 \ci{[0.174,0.206]} & $+0.026$ / $+11.0\,\text{px}$ & $0.128$ / $0.0080$ \\
CE head + Kabsch & 0.021 \ci{[0.017,0.026]} & 70.2 \ci{[66.9,73.6]} & 11.70 \ci{[11.27,12.12]} & \textbf{0.00 \ci{[0.00,0.00]}} & 0.214 \ci{[0.198,0.231]} & $+0.022$ / $+11.5\,\text{px}$ & $1.000$ / $0.0080$ \\
CE head (VFlash-style) & 0.021 \ci{[0.017,0.026]} & 70.5 \ci{[67.4,73.8]} & 12.38 \ci{[11.94,12.82]} & 5.40 \ci{[5.22,5.56]} & 0.214 \ci{[0.198,0.231]} & $+0.021$ / $+13.2\,\text{px}$ & $1.000$ / $0.0080$ \\
MDN head + Kabsch & 0.021 \ci{[0.017,0.025]} & 69.7 \ci{[66.4,73.0]} & 12.92 \ci{[12.47,13.35]} & \textbf{0.00 \ci{[0.00,0.00]}} & \textbf{0.174 \ci{[0.160,0.190]}} & $+0.025$ / $+11.6\,\text{px}$ & $1.000$ / $0.0080$ \\
MDN head (4 branches) & 0.021 \ci{[0.017,0.026]} & 69.5 \ci{[66.2,72.9]} & 13.09 \ci{[12.64,13.52]} & 2.63 \ci{[2.55,2.70]} & 0.174 \ci{[0.160,0.190]} & $+0.023$ / $+11.9\,\text{px}$ & $1.000$ / $0.0080$ \\
Bi-FPD + Kabsch & 0.025 \ci{[0.021,0.030]} & 66.7 \ci{[63.4,70.1]} & 14.28 \ci{[13.80,14.75]} & \textbf{0.00 \ci{[0.00,0.00]}} & 0.230 \ci{[0.211,0.250]} & $+0.023$ / $+10.0\,\text{px}$ & $0.128$ / $0.0080$ \\
Bi-FPD (mode-routed MDN) & 0.025 \ci{[0.021,0.030]} & 67.0 \ci{[63.7,70.3]} & 14.40 \ci{[13.92,14.86]} & 2.19 \ci{[2.12,2.25]} & 0.230 \ci{[0.211,0.250]} & $+0.023$ / $+9.9\,\text{px}$ & $0.168$ / $0.0080$ \\
Autoregressive (KV cache) & 0.059 \ci{[0.052,0.066]} & 50.7 \ci{[48.1,53.4]} & 28.32 \ci{[27.58,29.03]} & 3.83 \ci{[3.67,3.99]} & 0.247 \ci{[0.229,0.265]} & $+0.017$ / $+8.9\,\text{px}$ & $0.008$ / $0.0080$ \\
Procrustes-FPD, E2E-L1 + warped RGB & 0.028 \ci{[0.024,0.032]} & 68.2 \ci{[65.0,71.5]} & 13.78 \ci{[13.31,14.26]} & \textbf{0.00 \ci{[0.00,0.00]}} & 0.257 \ci{[0.237,0.278]} & $+0.020$ / $+6.9\,\text{px}$ & $0.008$ / $0.0080$ \\
Procrustes-FPD, Pre-L1 + warped RGB & 0.030 \ci{[0.025,0.035]} & 66.1 \ci{[62.8,69.4]} & 14.45 \ci{[13.97,14.95]} & \textbf{0.00 \ci{[0.00,0.00]}} & 0.260 \ci{[0.241,0.282]} & $+0.010$ / $+5.5\,\text{px}$ & $0.008$ / $0.0080$ \\
Procrustes-FPD, Pre-L1 + unwarped RGB & 0.029 \ci{[0.024,0.034]} & 68.3 \ci{[65.2,71.5]} & 13.71 \ci{[13.24,14.20]} & \textbf{0.00 \ci{[0.00,0.00]}} & 0.274 \ci{[0.253,0.294]} & $+0.011$ / $+6.3\,\text{px}$ & $0.008$ / $0.0080$ \\
\cmidrule(l){1-8}
WF control, no RGB: Procrustes-FPD, Pre-L1 & 0.043 \ci{[0.037,0.050]} & 59.1 \ci{[55.8,62.6]} & 19.31 \ci{[18.79,19.84]} & \textbf{0.00 \ci{[0.00,0.00]}} & 0.320 \ci{[0.298,0.344]} & $-0.003$ / $+0.7\,\text{px}$ & $0.008$ / $0.0080$ \\
\cmidrule(l){1-8}
WF control, no RGB: Soft-L1 head & 0.042 \ci{[0.036,0.049]} & 60.4 \ci{[57.2,63.7]} & 20.52 \ci{[20.01,21.04]} & 10.46 \ci{[10.15,10.77]} & 0.320 \ci{[0.298,0.344]} & $+0.000$ / $+0.4\,\text{px}$ & $0.008$ / $0.0080$ \\
\midrule
\textbf{\texttt{lerobot/pusht}, OF (3 seeds)} & \textbf{MAP-ADE Raw $\downarrow$} & \textbf{MAP-ADE Kabsch $\downarrow$} & \textbf{RB-Err Raw $\downarrow$} & \textbf{U-MAE Kabsch $\downarrow$} & \textbf{Brier $\downarrow$} & \textbf{Gain $\Delta_{\text{WF}-\text{OF}}$ (Raw / Kab)} & \textbf{$p_{\text{Holm}(10)}$ (Raw / Kab)} \\
\midrule
SE(2) tokens, Soft-L1 & \textbf{3.68 \ci{[3.38,4.00]}} & \textbf{3.68 \ci{[3.38,4.00]}} & \textbf{0.00 \ci{[0.00,0.00]}} & \textbf{0.021 \ci{[0.018,0.025]}} & 0.220 \ci{[0.190,0.250]} & $+9.46$ / $+9.46\,\text{px}$ & $0.0050$ / $1.0000$ \\
Soft-L1 head & 3.92 \ci{[3.65,4.21]} & \textbf{3.71 \ci{[3.44,4.00]}} & 1.43 \ci{[1.34,1.51]} & 0.022 \ci{[0.019,0.025]} & \textbf{0.205 \ci{[0.177,0.233]}} & $+8.86$ / $+8.05\,\text{px}$ & \textbf{Reference} \\
Procrustes-FPD, E2E-L1 + rotated RGB & 3.72 \ci{[3.47,3.98]} & 3.72 \ci{[3.47,3.98]} & \textbf{0.00 \ci{[0.00,0.00]}} & 0.022 \ci{[0.019,0.025]} & 0.220 \ci{[0.197,0.244]} & $+6.45$ / $+6.45\,\text{px}$ & $0.0320$ / $1.0000$ \\
Procrustes-FPD, E2E-L1 & 3.78 \ci{[3.52,4.05]} & \textbf{3.78 \ci{[3.52,4.05]}} & \textbf{0.00 \ci{[0.00,0.00]}} & 0.024 \ci{[0.021,0.027]} & 0.227 \ci{[0.196,0.258]} & $+6.39$ / $+6.39\,\text{px}$ & $0.0050$ / $0.3960$ \\
Soft-L1 head + rotated RGB & 5.70 \ci{[5.39,6.02]} & 5.47 \ci{[5.17,5.79]} & 1.66 \ci{[1.57,1.75]} & 0.031 \ci{[0.026,0.035]} & 0.272 \ci{[0.246,0.295]} & $+7.09$ / $+6.29\,\text{px}$ & $0.0050$ / $0.0050$ \\
CE head (VFlash-style) & 4.29 \ci{[3.95,4.67]} & 4.13 \ci{[3.79,4.50]} & 1.14 \ci{[1.02,1.27]} & 0.024 \ci{[0.020,0.027]} & 0.223 \ci{[0.194,0.251]} & $+12.24$ / $+10.25\,\text{px}$ & $0.0050$ / $0.0050$ \\
MDN head (4 branches) & 5.25 \ci{[4.86,5.68]} & 5.19 \ci{[4.80,5.62]} & 0.67 \ci{[0.61,0.73]} & 0.026 \ci{[0.022,0.030]} & 0.236 \ci{[0.201,0.271]} & $+11.10$ / $+10.42\,\text{px}$ & $0.0050$ / $0.0050$ \\
Bi-FPD (mode-routed MDN) & 5.32 \ci{[4.87,5.79]} & 5.27 \ci{[4.82,5.73]} & 0.59 \ci{[0.54,0.64]} & 0.026 \ci{[0.022,0.029]} & 0.228 \ci{[0.198,0.259]} & $+11.43$ / $+10.52\,\text{px}$ & $0.0050$ / $0.0050$ \\
Autoregressive (KV cache) & 5.58 \ci{[5.17,6.03]} & 5.43 \ci{[5.03,5.87]} & 1.05 \ci{[0.96,1.15]} & 0.031 \ci{[0.027,0.035]} & 0.233 \ci{[0.200,0.266]} & $+19.85$ / $+18.69\,\text{px}$ & $0.0050$ / $0.0050$ \\
\cmidrule(l){1-8}
WF control, no RGB: Soft-L1 head & 13.25 \ci{[12.42,14.13]} & 11.84 \ci{[11.04,12.72]} & 7.90 \ci{[7.43,8.38]} & 0.069 \ci{[0.060,0.078]} & 0.277 \ci{[0.242,0.313]} & $-0.46$ / $-0.08\,\text{px}$ & $0.0050$ / $0.0050$ \\
\bottomrule
\end{tabular}%
}
\end{table*}

\subsection{Registered Frame \texorpdfstring{$\times$}{x} Decoder Grid, Diffusion Correction and Data Scaling}
\label{sec:grid_scaling}
\begin{figure*}[t]
\centering
\includegraphics[width=0.94\textwidth]{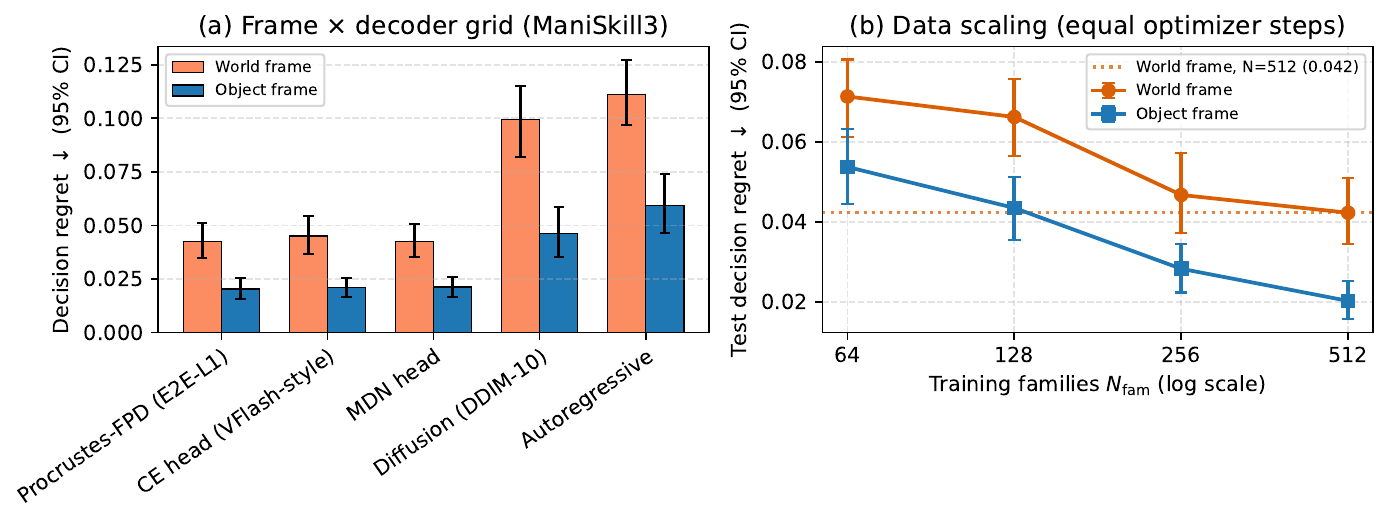}
\caption{Registered frame $\times$ decoder grid and data scaling on ManiSkill3 (hierarchical seed $\times$ family 95\% CIs; no RGB input in either frame). \textbf{(a)} The object frame lowers regret by $1.9\times$--$2.2\times$ for all five decoders; in either frame, the iterative decoders (diffusion, autoregression) have $2.3\times$--$2.9\times$ the regret of the single-pass ones. \textbf{(b)} With equal optimizer steps, the object-frame model trained on $128$ families matches the world-frame model trained on $512$.}
\label{fig:v4_grid_scaling}
\end{figure*}

The comparisons above mix models from different stages, and the object-frame models differ from their world-frame counterparts in their RGB input. We therefore registered a $2{\times}5$ grid in which all decoders share one training path, no model receives RGB, and only the frame and the decoder vary (Table~\ref{tab:v4_grid_scaling}, Fig.~\ref{fig:v4_grid_scaling}). Its intervals come from a seed $\times$ family hierarchical bootstrap, which includes training-seed variation.

\textbf{Diffusion baseline correction.} While preparing the grid we found an error in our diffusion baseline. The model class defined a static method \texttt{norm} that shadowed the module's final RMSNorm layer of the same name, so the original diffusion model ran without its final normalization. Its fixed target scaling also clipped object-frame targets at $\pm 1.2$. The corrected model restores the normalization and fits the target mean and $99.9$th-percentile scale on the training split, in whichever frame the model is trained. The original numbers are superseded: we report them only in Appendix~\ref{app:superseded} (Table~\ref{tab:superseded}). They are excluded from the other tables and figures, but they remain in the Holm families, which is conservative. With the correction, the object frame lowers diffusion regret on ManiSkill3 by $2.16\times$ ($0.0993 \to 0.0460$, $\pholm{9}{=}0.0045$), MAP-ADE by $2.15\times$ ($59.73 \to 27.84\,\text{px}$), and projected MAP-ADE on \texttt{lerobot/pusht} by $2.99\times$ ($55.19 \to 18.45\,\text{px}$).

\textbf{H1 and H2: frame versus decoder.} The object frame significantly lowers regret and MAP-ADE for all five decoders (Procrustes-FPD E2E-L1, CE head, MDN head, diffusion and autoregression; $\pholm{9}{<}0.005$), so H1 is supported. H2 compared the mean frame effect $F$ (world-frame minus object-frame regret, averaged over decoders) with the decoder spread $S$ (worst minus best decoder in the object frame). For regret, $F = 0.0345$ $[0.0283, 0.0415]$ and $S = 0.0389$ $[0.0294, 0.0551]$, so $F - S = -0.0044$ $[-0.0226, +0.0074]$. For MAP-ADE on ManiSkill3, $F = 17.35$ and $S = 16.97\,\text{px}$, so $F - S = +0.37$ $[-5.67, +3.24]\,\text{px}$; on \texttt{lerobot/pusht} the values are $16.55$ and $14.67\,\text{px}$. The registered criterion, a lower confidence bound of $F - S$ above zero, is not met: the two effects are statistically comparable. The spread comes from the iterative decoders, whose object-frame regret is $2.3\times$ (diffusion, $0.0460$) and $2.9\times$ (autoregression, $0.0592$) that of Procrustes-FPD. In an analysis that was not registered, the three single-pass heads are statistically indistinguishable on regret ($0.0203$, $0.0209$, $0.0212$; each against Procrustes-FPD, $p{>}0.69$), while the frame changes each of them by $0.021$--$0.024$ (frame effect minus single-pass spread: $0.0215$ $[0.0130, 0.0271]$). We read the grid as two separable effects of similar size: the coordinate frame (about $2\times$ for every decoder) and single-pass versus iterative decoding (about $2\times$--$3\times$). The choice of single-pass head is secondary.

\textbf{H3 and H4: single-pass decoding against a DDIM step sweep.} In the object frame, Procrustes-FPD (E2E-L1) reaches $0.0203$ regret and $11.35\,\text{px}$ MAP-ADE at $1.62\,\text{ms}$ per decision. It ties the CE head ($0.0209$) and the MDN head ($0.0212$) on regret ($\pholm{9}{=}1.0$) and has lower MAP-ADE than all four other decoders ($\pholm{9}{<}0.005$). We ran the corrected object-frame diffusion model with $1$, $2$, $4$, $10$, $25$ and $50$ DDIM steps (Table~\ref{tab:v4_grid_scaling}, bottom right). With $1$ or $2$ steps, sampling from this $\epsilon$-prediction model fails ($0.1315$ regret, MAP-ADE above $330\,\text{px}$), as expected when a single update must move from almost pure noise to the data. From $4$ steps on, diffusion reaches $0.0558$ regret at $4.78\,\text{ms}$, $0.0460$ at $10.32\,\text{ms}$, and at best $0.0400$ at $47.02\,\text{ms}$. Procrustes-FPD is both more accurate and faster than every setting, with $3\times$--$29\times$ lower latency than the $4$- to $50$-step settings, so H4 is supported. We did not test diffusion models trained or distilled for one or two steps, such as consistency models.

\textbf{H5: data scaling.} We trained Procrustes-FPD (E2E-L1) on nested subsets of $N_{\text{fam}} \in \{64, 128, 256, 512\}$ training families with the same number of optimizer steps (Table~\ref{tab:v4_grid_scaling}, bottom left; Fig.~\ref{fig:v4_grid_scaling}b). The object-frame model trained on $128$ families ($0.0435$ regret, $19.37\,\text{px}$) matches the world-frame model trained on $512$ ($0.0423$ regret, $19.11\,\text{px}$), roughly a $4\times$ gain in sample efficiency. The gap between frames shrinks with data on MAP-ADE ($18.39$, $13.20$, $10.26$, $7.76\,\text{px}$) but not on regret ($+0.0176$, $+0.0228$, $+0.0185$, $+0.0220$). H5 is therefore only partly supported: the object frame improves data efficiency, and at $512$ families its regret advantage has not yet narrowed.

\begin{table*}[t]
\centering
\caption{Registered $2{\times}5$ coordinate-frame $\times$ decoder grid, DDIM step sweep and data-scaling study (seed $\times$ family/episode hierarchical bootstrap, $B{=}2000$, 95\% CIs). All ten grid models share one training path and receive no RGB input, so only the coordinate frame (WF: world frame; OF: object frame) and the decoder change. The diffusion model is the corrected version (Section~\ref{sec:grid_scaling}). $p_{\text{Holm}(9)}$: Holm correction over the nine registered comparisons per metric. Latency: median over 200 calls for one decision (batch of 4 candidate actions, $S{=}16$ samples where applicable), measured in the grid run.}
\label{tab:v4_grid_scaling}
\resizebox{\textwidth}{!}{%
\begin{tabular}{lccccccc}
\toprule
\textbf{Decoder} & \textbf{ManiSkill3 WF Regret} & \textbf{ManiSkill3 OF Regret} & \textbf{Frame Gain $\Delta_{\text{WF}-\text{OF}}$ ($p_{\text{Holm}(9)}$)} & \textbf{ManiSkill3 WF $\to$ OF ADE (px)} & \textbf{vs.\ OF Procrustes-FPD: Reg / ADE $p_{\text{Holm}(9)}$} & \textbf{PushT WF $\to$ OF ADE (px)} & \textbf{Latency (ms)} \\
\midrule
\textbf{Procrustes-FPD, E2E-L1 (single pass)} & 0.0423 \ci{[0.0345,0.0511]} & \textbf{0.0203} \ci{[0.0157,0.0253]} & $\mathbf{+0.0220}$ ($0.0045$) & 19.1 $\to$ \textbf{11.3} & \textbf{Reference} & 12.00 $\to$ \textbf{3.78} & 1.62 \\
CE head, VFlash-style (single pass) & 0.0450 \ci{[0.0367,0.0544]} & 0.0209 \ci{[0.0167,0.0255]} & $\mathbf{+0.0241}$ ($0.0045$) & 25.2 $\to$ \textbf{12.4} & $1.000$ / $0.0045$ & 13.66 $\to$ \textbf{4.13} & 1.76 \\
MDN head, 4 branches (single pass) & 0.0425 \ci{[0.0352,0.0507]} & 0.0212 \ci{[0.0167,0.0260]} & $\mathbf{+0.0213}$ ($0.0045$) & 21.0 $\to$ \textbf{13.1} & $1.000$ / $0.0045$ & 14.74 $\to$ \textbf{5.19} & 1.68 \\
Diffusion, DDIM-10, corrected ($S{=}16$) & 0.0993 \ci{[0.0817,0.1153]} & 0.0460 \ci{[0.0353,0.0584]} & $\mathbf{+0.0532}$ ($0.0045$) & 59.7 $\to$ \textbf{27.8} & $0.005$ / $0.0045$ & 55.19 $\to$ \textbf{18.45} & 10.32 \\
Autoregressive, KV cache ($S{=}16$) & 0.1110 \ci{[0.0968,0.1272]} & 0.0592 \ci{[0.0466,0.0738]} & $\mathbf{+0.0518}$ ($0.0045$) & 54.7 $\to$ \textbf{28.3} & $0.005$ / $0.0045$ & 24.12 $\to$ \textbf{5.43} & 12.89 \\
\midrule
\multicolumn{8}{l}{\textbf{H2 (ManiSkill3):} mean frame effect $F = 0.0345$ regret ($17.35\,\text{px}$ ADE); five-decoder spread $S = 0.0389$ regret ($16.97\,\text{px}$ ADE); $F - S = -0.0044$ $[-0.0226, +0.0074]$ regret, $+0.37$ $[-5.67, +3.24]\,\text{px}$ ADE. Registered criterion ($F - S > 0$) not met.} \\
\midrule
\multicolumn{4}{l}{\textbf{H5: data scaling (Procrustes-FPD, E2E-L1; equal optimizer steps)}} & \multicolumn{4}{l}{\textbf{H4: DDIM step sweep (corrected diffusion, OF)}} \\
\cmidrule(r){1-4}\cmidrule(l){5-8}
\textbf{Train Families $N_{\text{fam}}$} & \textbf{WF Regret / ADE} & \textbf{OF Regret / ADE} & \textbf{Frame Gain $\Delta$ Reg / ADE} & \textbf{DDIM Steps} & \textbf{OF Regret $\downarrow$} & \textbf{OF MAP-ADE $\downarrow$ (px)} & \textbf{Latency (ms)} \\
\midrule
$N = 64$ & 0.0714 / 40.9\,px & \textbf{0.0538} / \textbf{22.5\,px} & $+0.0176$ / $+18.4\,\text{px}$ & 4 steps & 0.0558 \ci{[0.0412,0.0725]} & 32.93 & 4.78 \\
$N = 128$ & 0.0663 / 32.6\,px & \textbf{0.0435} / \textbf{19.4\,px} & $+0.0228$ / $+13.2\,\text{px}$ & 10 steps & 0.0460 \ci{[0.0353,0.0584]} & 27.84 & 10.32 \\
$N = 256$ & 0.0468 / 23.7\,px & \textbf{0.0283} / \textbf{13.5\,px} & $+0.0185$ / $+10.3\,\text{px}$ & 25 steps & 0.0419 \ci{[0.0324,0.0534]} & 25.98 & 24.24 \\
$N = 512$ & 0.0423 / 19.1\,px & \textbf{0.0203} / \textbf{11.3\,px} & $+0.0220$ / $+7.8\,\text{px}$ & 50 steps & 0.0400 \ci{[0.0316,0.0496]} & 24.77 & 47.02 \\
\bottomrule
\end{tabular}%
}
\end{table*}

\section{Limitations}
\label{sec:limitations}
\begin{itemize}
\item \textbf{Scope.} We study one task family, planar pushing of a single rigid object, with $2$M- to $4$M-parameter models. Whether the frame effect carries over to several objects, articulated or deformable objects, or full 3D pose is untested, as is its size for large pretrained backbones.
\item \textbf{Simulation only.} Both benchmarks are simulated. \texttt{lerobot/pusht} contains human demonstrations, but they were recorded in a 2D simulator. We do not evaluate on a physical robot, and we evaluate the world model by decision regret and trajectory error, not by closed-loop task success.
\item \textbf{Reuse of test data.} The ManiSkill3 test split was sealed for the pre-registered benchmark, but the exploratory controls and the registered grid were evaluated on the same test families after it was unsealed. The object-frame result was found in exploratory runs and then replicated in the registered grid. It was not predicted before the test data were seen.
\item \textbf{Run-to-run variation.} Identical retraining moved regret by up to $0.0035$. Tables~\ref{tab:maniskill_main}--\ref{tab:se2_canonical} use family-cluster intervals that do not include this variation; only the grid (Table~\ref{tab:v4_grid_scaling}) resamples seeds.
\item \textbf{Baselines.} Our iterative baselines are small models trained from scratch in our code base, and the diffusion baseline needed a correction. The corrected diffusion model inherits the hyperparameters selected for the original one, without re-tuning. Re-tuned, stronger or distilled diffusion models may narrow the gap. The VFlash-style head reimplements VFlash's loss and decoding in our backbone; it is not the released VFlash model.
\end{itemize}

\section{Conclusion}
We studied single-pass slot decoders as System-1 world models, i.e.\ fast action-conditioned predictors, using a pre-registered benchmark, matched exploratory controls and a registered $68$-run frame $\times$ decoder and data-scaling grid on ManiSkill3 \texttt{PushT-v1} and \texttt{lerobot/pusht}. Two factors largely determine how well such a predictor ranks candidate actions. The first is the coordinate frame: expressing inputs and outputs in the $\SE$ frame of the object's initial pose lowers test regret by about $2\times$ for every decoder we tested and gives roughly $4\times$ sample efficiency. The second is single-pass versus iterative decoding: in the object frame, single-pass heads have $2.3\times$--$2.9\times$ lower regret than corrected DDIM diffusion and KV-cached autoregression and are several times faster, while the choice among single-pass heads has no measurable effect on regret. Our registered hypothesis that the frame effect exceeds the decoder spread was not supported; the two are of comparable size. Soft-token expectation with an end-to-end 2D Kabsch projection gives the lowest trajectory error, rigid-body consistency by construction, and lower regret than diffusion at every budget from $4$ to $50$ DDIM steps. For practitioners, the recommendation is to choose the coordinate frame first and then decode in a single pass. Open questions are whether the frame effect holds for tasks with richer symmetry, such as several objects or full 3D pose, and how diffusion models distilled for one or two steps compare.

\section*{Acknowledgment}
AI assistants were used to help write and debug the experiment code, to run and audit the experiments, and to draft and edit the text. The author designed the study, checked the results and the manuscript, and takes full responsibility for the content.

\textit{Availability.} Code, frozen configurations, per-seed raw outputs and the script that generates every table and figure in this paper will be released publicly.

\begin{table*}[!t]
\centering
\footnotesize
\caption{Model names used in this paper and the corresponding identifiers in the released code and result files. Pre-reg.: pre-registered benchmark arm; Expl.: exploratory control; Grid: registered frame $\times$ decoder grid, where every decoder is run as \texttt{WF\_NoRGB\_}$\langle$id$\rangle$ and \texttt{BF\_}$\langle$id$\rangle$.}
\label{tab:names}
\begin{tabular}{llllc}
\toprule
\textbf{Name} & \textbf{Frame} & \textbf{Identifier} & \textbf{Status} & \textbf{Tables} \\
\midrule
Procrustes-FPD, Pre-L1 & WF & \texttt{A3\_VFlashSoftL1\_Kabsch} & Expl. & I--III \\
Soft-L1 head & WF & \texttt{A3\_VFlashSoftL1} & Pre-reg. & I--III \\
Procrustes-FPD, E2E-L1 & WF & \texttt{A3\_VFlashSoftL1\_E2EKabsch} & Expl. & I--III \\
SE(2) tokens, Soft-L1 & WF & \texttt{SE2\_VFlashSoftL1} & Expl. & I--III \\
CE head (VFlash-style) & WF & \texttt{A2\_VFlashPureCE} & Pre-reg. & I--III \\
CE head + Kabsch & WF & \texttt{A2\_VFlashPureCE\_Kabsch} & Expl. & I--III \\
SE(2) tokens, CE & WF & \texttt{SE2\_VFlashPureCE} & Expl. & I--III \\
MDN head (4 branches) & WF & \texttt{A6\_GenericMDN} & Pre-reg. & I--III \\
MDN head + Kabsch & WF & \texttt{A6\_GenericMDN\_Kabsch} & Expl. & I--III \\
SE(2) tokens, MDN & WF & \texttt{SE2\_GenericMDN} & Expl. & I--III \\
Bi-FPD (mode-routed MDN) & WF & \texttt{A12\_BiFPD} & Pre-reg. & I--III \\
Direct utility regression & WF & \texttt{A1\_DirectScore} & Pre-reg. & I--III \\
Contact-mode classifier & WF & \texttt{A8\_ClassifierOnlyTuned} & Pre-reg. & I--III \\
Autoregressive (KV cache) & WF & \texttt{A4\_ARSeqSampled} & Pre-reg. & I--III \\
Diffusion, DDIM-10 (superseded) & WF & \texttt{A5\_DiffusionTraj} & Pre-reg. & I--III, VII \\
Procrustes-FPD, E2E-L1 & OF & \texttt{BF\_A3\_VFlashSoftL1\_E2EKabsch} & Expl. & IV \\
Procrustes-FPD, Pre-L1 & OF & \texttt{BF\_Equiv\_ProcrustesFPD} & Expl. & IV \\
Soft-L1 head & OF & \texttt{BF\_Equiv\_RawSoftL1} & Expl. & IV \\
SE(2) tokens, Soft-L1 & OF & \texttt{BF\_SE2\_VFlashSoftL1} & Expl. & IV \\
CE head + Kabsch & OF & \texttt{BF\_A2\_VFlashPureCE\_Kabsch} & Expl. & IV \\
CE head (VFlash-style) & OF & \texttt{BF\_A2\_VFlashPureCE} & Expl. & IV \\
MDN head + Kabsch & OF & \texttt{BF\_A6\_GenericMDN\_Kabsch} & Expl. & IV \\
MDN head (4 branches) & OF & \texttt{BF\_A6\_GenericMDN} & Expl. & IV \\
Bi-FPD + Kabsch & OF & \texttt{BF\_A12\_BiFPD\_Kabsch} & Expl. & IV \\
Bi-FPD (mode-routed MDN) & OF & \texttt{BF\_A12\_BiFPD} & Expl. & IV \\
Autoregressive (KV cache) & OF & \texttt{BF\_A4\_ARSeqSampled} & Expl. & IV \\
Diffusion, DDIM-10 (superseded) & OF & \texttt{BF\_A5\_DiffusionTraj} & Expl. & IV, VII \\
Procrustes-FPD, E2E-L1 + warped RGB & OF & \texttt{BF\_RotRGB\_E2EKabsch} & Expl. & IV \\
Procrustes-FPD, Pre-L1 + warped RGB & OF & \texttt{BF\_RotRGB\_ProcrustesFPD} & Expl. & IV \\
Procrustes-FPD, Pre-L1 + unwarped RGB & OF & \texttt{BF\_WithRGB\_ProcrustesFPD} & Expl. & IV \\
WF control, no RGB: Procrustes-FPD, Pre-L1 & WF & \texttt{WF\_NoRGB\_ProcrustesFPD} & Expl. & IV \\
WF control, no RGB: Soft-L1 head & WF & \texttt{WF\_NoRGB\_RawSoftL1} & Expl. & IV \\
Soft-L1 head & OF & \texttt{BF\_A3\_VFlashSoftL1} & Expl. & IV \\
Procrustes-FPD, E2E-L1 + rotated RGB & OF & \texttt{BF\_OrthoRGB\_A3\_VFlashSoftL1\_E2EKabsch} & Expl. & IV \\
Soft-L1 head + rotated RGB & OF & \texttt{BF\_OrthoRGB\_A3\_VFlashSoftL1} & Expl. & IV \\
Diffusion, DDIM-10 (corrected) & WF / OF & \texttt{A5fix\_DiffusionTraj} & Grid & V, VII \\
\bottomrule
\end{tabular}
\end{table*}

\appendices
\section{Superseded Diffusion Baseline}
\label{app:superseded}
Table~\ref{tab:superseded} lists the results of the original diffusion baseline, whose final normalization layer was disabled (Section~\ref{sec:grid_scaling}), next to the corrected model. The two sets of rows use different protocols: the superseded world-frame rows had RGB input, the corrected rows come from the grid without RGB, and their intervals are computed differently. They should therefore not be compared row by row. The correction mainly changed the object-frame behaviour. The superseded model became worse in the object frame (regret $0.084 \to 0.091$), whereas the corrected model improves by $2.16\times$, in line with the other decoders.

\begin{table}[t]
\centering
\caption{Superseded diffusion baseline (final RMSNorm disabled by a bug) and the corrected model. Superseded rows follow the protocols of Tables~\ref{tab:maniskill_main}, \ref{tab:public_pusht} and~\ref{tab:se2_canonical} (world-frame rows with RGB input); corrected rows are from the registered grid (no RGB input, hierarchical bootstrap; Table~\ref{tab:v4_grid_scaling}). \texttt{lerobot/pusht} MAP-ADE is after Kabsch projection. 10 DDIM steps, $S{=}16$.}
\label{tab:superseded}
\resizebox{\columnwidth}{!}{%
\begin{tabular}{llcc}
\toprule
\textbf{Model} & \textbf{Setting} & \textbf{Regret $\downarrow$} & \textbf{MAP-ADE $\downarrow$ (px)} \\
\midrule
Superseded & ManiSkill3, WF & 0.084 \ci{[0.075,0.095]} & 29.9 \ci{[29.4,30.4]} \\
Superseded & ManiSkill3, OF & 0.091 \ci{[0.082,0.102]} & 38.6 \ci{[38.2,39.0]} \\
Superseded & \texttt{lerobot/pusht}, WF & $-$ & 14.78 \ci{[14.16,15.43]} \\
Superseded & \texttt{lerobot/pusht}, OF & $-$ & 18.26 \ci{[17.86,18.68]} \\
\midrule
Corrected & ManiSkill3, WF & 0.0993 \ci{[0.0817,0.1153]} & 59.7 \\
Corrected & ManiSkill3, OF & 0.0460 \ci{[0.0353,0.0584]} & 27.8 \\
Corrected & \texttt{lerobot/pusht}, WF & $-$ & 55.19 \\
Corrected & \texttt{lerobot/pusht}, OF & $-$ & 18.45 \\
\bottomrule
\end{tabular}%
}
\end{table}

\section{Model Names and Code Identifiers}
\label{app:names}
Table~\ref{tab:names} maps the model names used in the text to the identifiers used in the released code and result files.

\end{document}